\documentclass[12pt]{article}

\usepackage{graphicx}
\graphicspath{{./images/}}
\usepackage{floatpag} 
\usepackage{gensymb} 

\usepackage{tabularx}
\usepackage{multirow}
\usepackage{threeparttable}
\usepackage{booktabs}
\usepackage{caption}
\usepackage[table,xcdraw]{xcolor}

\usepackage{amssymb}
\usepackage{pifont}
\newcommand{\cmark}{\ding{51}}%
\newcommand{\xmark}{\ding{55}}%

\newcommand{\beginsupplement}{%
    \setcounter{table}{0}
    \renewcommand{\thetable}{S\arabic{table}}%
    \setcounter{figure}{0}
    \renewcommand{\thefigure}{S\arabic{figure}}%
 }

\usepackage{times}
\usepackage{setspace}
\usepackage[hidelinks,colorlinks=true,linkcolor=blue,citecolor=blue,urlcolor=blue]{hyperref}

\newenvironment{sciabstract}{%
\begin{quote} \bf}
{\end{quote}}

\title{Navigating to Objects in the Real World}

\author
{Theophile Gervet,$^{1\ast}$ Soumith Chintala,$^{4}$ \\
Dhruv Batra,$^{3,4}$ Jitendra Malik,$^{2,4}$ Devendra Singh Chaplot$^{4}$\\
\\
\normalsize{$^{1}$Carnegie Mellon University,}\\
\normalsize{$^{2}$University of California, Berkeley,}\\
\normalsize{$^{3}$Georgia Institute of Technology,}\\
\normalsize{$^{4}$Meta AI Research}\\
\normalsize{$^\ast$To whom correspondence should be addressed; E-mail:  tgervet@andrew.cmu.edu.}\\
\normalsize{\href{https://theophilegervet.github.io/projects/real-world-object-navigation}{Project Website}}
}

\date{}

\begin{document} 

\baselineskip24pt
\captionsetup[figure]{labelfont={bf},name={Fig.}, labelsep=period}
\captionsetup[table]{labelfont={bf},name={Table}, labelsep=period}
\maketitle

{\setstretch{1.0} 
\begin{sciabstract} 
Semantic navigation is necessary to deploy mobile robots in uncontrolled environments like our homes, schools, and hospitals. 
Many learning-based approaches have been proposed in response to the lack of semantic understanding of the classical pipeline for spatial navigation, which builds a geometric map using depth sensors and plans to reach point goals.
Broadly, end-to-end learning approaches reactively map sensor inputs to actions with deep neural networks, while modular learning approaches enrich the classical pipeline with learning-based semantic sensing and exploration.
But learned visual navigation policies have predominantly been evaluated in simulation. 
How well do different classes of methods work on a robot? 
We present a large-scale empirical study of semantic visual navigation methods comparing representative methods from classical, modular, and end-to-end learning approaches across six homes with no prior experience, maps, or instrumentation.
We find that modular learning works well in the real world, attaining a $90$\% success rate. 
In contrast, end-to-end learning does not, dropping from $77$\% simulation to $23$\% real-world success rate due to a large image domain gap between simulation and reality. 
For practitioners, we show that modular learning is a reliable approach to navigate to objects: modularity and abstraction in policy design enable Sim-to-Real transfer.
For researchers, we identify two key issues that prevent today's simulators from being reliable evaluation benchmarks — (A) a large Sim-to-Real gap in images and (B) a disconnect between simulation and real-world error modes — and propose concrete steps forward.
\end{sciabstract}

\begin{figure}
    \thisfloatpagestyle{empty}
    \centering
    \includegraphics[height=18cm,keepaspectratio]{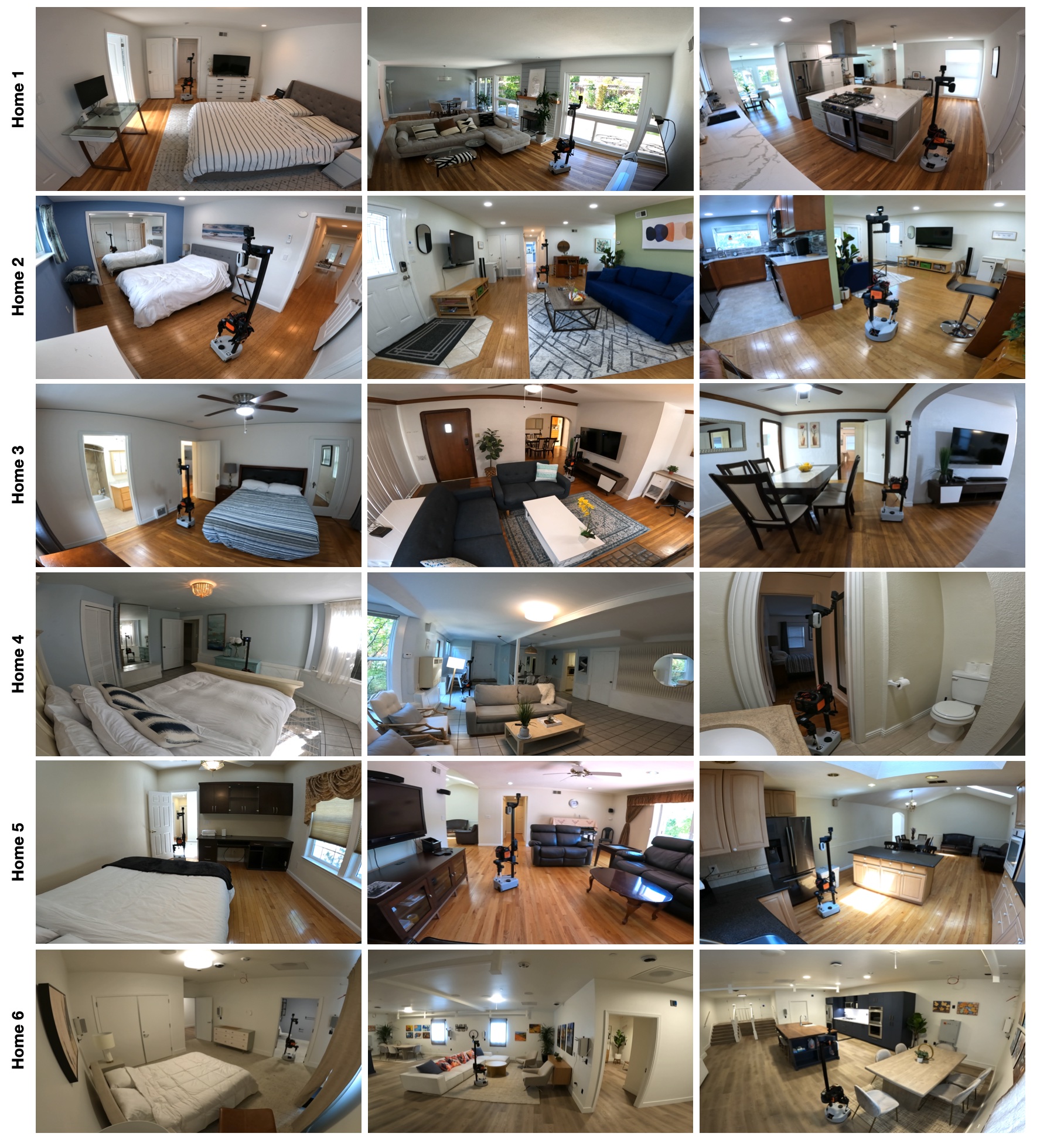}
    \caption{\textbf{Deployment of the semantic navigation policies in six visually diverse homes.}}
    \label{fig:home_diversity}
\end{figure}

\section{Introduction} 

Humans can navigate in unseen environments effortlessly. We can utilize our experience in prior environments to explore any new environment and find any target object efficiently. For example, when looking for a glass of water at a friend's house we're visiting for the first time, we can easily find the kitchen without going to bedrooms or storage closets. Learning such semantic priors is essential to deploy autonomous mobile robots in uncontrolled environments like our homes, schools, and hospitals. In this work, we tackle the Object Goal navigation~\cite{anderson2018evaluation} task where a robot is asked to find an object belonging to a particular category, like a bed or a couch, in a completely unseen environment. 

Navigation has been studied in robotics literature for over three decades~\cite{moravec1980obstacle, chatila1985position, elfes1987sonar, elfes1989using, kuipers1991robot, cox1991blanche, leonard1992dynamic, fox1997dynamic, burgard1999experiences, thrun1999minerva, thrun2001robust, thrun2002robotic}. Many classical approaches to navigation require access to pre-computed maps~\cite{cox1991blanche, burgard1999experiences, thrun1999minerva}. Among approaches that can operate in unseen environments, classical approaches typically build a geometric map of the environment using depth sensors~\cite{thrun2001robust, thrun2002robotic, newcombe2011kinectfusion}, and later a monocular RGB camera~\cite{davison2007monoslam, jones2011visual, sattler2018benchmarking}, while simultaneously localizing the robot relative to its growing map. Building on this simultaneous localization and mapping (SLAM) module, classical methods explore with heuristics such as frontier-based exploration~\cite{yamauchi1997frontier} and leverage an analytical planner for low-level control to avoid obstacles and reach exploration or point goals. Adapting these methods to navigate to objects requires detecting objects, keeping objects in memory, and exploring semantically towards objects. Semantic SLAM methods~\cite{flint2011manhattan, kundu2014joint, bowman2017probabilistic, ma2017multi, zhang2018semantic, rosinol2020kimera, salas2013slam++} naturally extend SLAM to detect objects and keep them in memory but offer no solution for efficient semantic exploration.

Following recent advances in machine learning and computer vision, there has been a lot of interest in designing learning-based policies for visual navigation capable of learning these semantic priors. The most common learning-based methods for semantic navigation use a deep neural network, usually consisting of a visual encoder followed by a recurrent layer for memory, to predict actions directly from raw observations. These \textit{end-to-end learning} approaches are trained using backpropagation with imitation learning (IL) or reinforcement learning (RL) losses. Inspired by seminal proofs of concept in autonomous driving ~\cite{pomerleau1988alvinn, muller2005off} and early successes of pixel-to-action deep reinforcement learning~\cite{mnih2015human, lillicrap2015continuous}, early applications of end-to-end learning to semantic navigation include~\cite{lample2016playing, zhu2017target, mirowski2016learning, dosovitskiy2016learning, chaplot2017arnold, savva2017minos, hermann2017grounded, chaplot2017gated, mirowski2018learning, codevilla2018end,banino2018vector}. Most relevant to us,~\cite{ye2021auxiliary, maksymets2021thda, ramrakhya2022habitat} propose end-to-end policies to navigate to object goals. End-to-end RL methods have been scaled to train with billions of frames corresponding to $80$ years of navigation time (in sim) using distributed training~\cite{wijmans2019decentralized} or tens of thousands of procedurally generated scenes~\cite{deitke2022procthor}. While end-to-end policies often directly map sensor data to actions, akin to a modern version of Rodney Brook's Subsumption architecture~\cite{brooks1986robust}, researchers have also introduced some structure into the neural network, such as intermediate spatial~\cite{gupta2017cognitive, parisotto2017neural, chaplot2018active, henriques2018mapnet, gordon2018iqa} and topological representations~\cite{yang2018visual, savinov2018semi, savinov2018episodic}.

Classical and end-to-end learning-based approaches offer distinct advantages. End-to-end learning policies can learn semantic priors for goal-directed exploration, while the modularity of the classical pipeline offers easier engineering and interpretability~\cite{mcallister2017concrete}. Following successful applications across robotics in autonomous driving~\cite{mcallister2017concrete, muller2018driving}, flight~\cite{scaramuzza2014vision}, and grasping~\cite{mousavian20196, mahler2017dex, morrison2018cartman}, much work has focused on designing \textit{modular learning} approaches that aim to combine the benefits of both learning and classical methods.
Modular learning approaches preserve the structure of the classical pipeline and replace analytical modules for specific subtasks with learned ones. In semantic navigation, the subtask decomposition typically includes separate modules for perception (object detection, mapping, pose estimation, SLAM), encoding goals, global waypoint selection policies, planning, and local obstacle avoidance policies. Learned modules are trained using direct supervision, which offers better sample efficiency than end-to-end learning~\cite{mcallister2017concrete}. Modular learning can enable Sim-to-Real transfer~\cite{muller2018driving, chaplot2020learning, chaplot2020object}: one can shield learned modules from Sim-to-Real domain gaps by designing abstractions of the input raw sensor data that contain sufficient information to solve the task while being invariant to environmental factors that are hard to simulate accurately (like photo-realistic RGB images) and thus unlikely to transfer from sim to reality. Representative examples of modular learning for exploration~\cite{chaplot2020learning, ramakrishnan2020occupancy} and for reaching object goals and image goals are~\cite{chaplot2020object, chaplot2020neural, ramakrishnan2022poni, hahn_nrns_2021}. Most relevant to us,~\cite{chaplot2020object} cleanly isolates the problem of learning a policy to explore semantically towards objects from the rest of the navigation problem. Modular learning methods have also been applied to longer-horizon tasks such as following language navigation instructions~\cite{Krantz_2021_ICCV, an20221st}, executing language instructions interactively in ALFRED~\cite{min2021film, liu9planning, murray2022following}, object rearrangement in AI2 Thor~\cite{sarch2022tidee, trabucco2022simple} and improving perception using active exploration~\cite{chaplot2020semantic, chaplot2021seal}.

Over the past few years, the semantic navigation community has proposed hundreds of methods and organized tens of benchmarks in sim~\cite{duan2022survey}. If the field doesn't lack proposed methods, what is missing to enable robots to navigate semantically? In our view, the missing piece of the puzzle is large-scale real-world evaluation. Learned navigation policies have predominantly been evaluated in sim~\cite{mishkin2019benchmarking}. We can attribute this to (A) the emergence of sophisticated embodied simulators~\cite{habitat19iccv,ai2thor,savva2017minos}, which significantly reduced the field's barrier to entry and sped up the proposal of new methods, and (B) the relative operational difficulties of bringing a robot out of the lab into diverse, realistic, deployment environments. But while simulators can be very useful for training, they are insufficient for evaluation. In the end, how well a method performs on a robot is the only thing that matters, and we have only partial answers, if any, as to whether sim is a good evaluation benchmark for semantic navigation. Does sim performance reflect real-world performance? Do design choices that improve sim performance improve real-world performance? Do sim error modes reflect real-world error modes? Most prior Sim-to-Real studies in navigation focused on spatial (point goal) navigation and legged locomotion~\cite{kadian2020sim2real, truong2021bi, truong2022rethinking, fu2022coupling, miki2022learning}, as opposed to semantic navigation. The few other real-world semantic navigation works directly train on real-world images for outdoor visual goals~\cite{shah2021ving, shah2022viking} and language instruction following~\cite{anderson2018vision}. A lack of real-world evaluation opens semantic navigation to the risk of sim-only research that does not generalize to the real world~\cite{hofer2021sim2real}.

Our proposed work addresses this issue through a large-scale empirical evaluation of semantic navigation policies. We compare representative methods from classical, modular learning, and end-to-end learning approaches across six visually diverse real home environments, as illustrated in Fig.~\ref{fig:home_diversity}. This represents $45$ hours of robot experiments ($3$ methods x $6$ homes x $10$ episodes per home x $15$ minutes per episode). In addition, we replicate one home in sim to ensure our experimental setting matches that of sim benchmarks and decompose the Sim-to-Real performance gap. We find that modular learning transfers to the real world very well, with performance rising from a $81$\% success rate in sim to $90$\% in the real world (within a limited time budget). In contrast, end-to-end learning performance drops sharply from $77$\% sim to $23$\% real-world success rate due to a large image domain gap between sim and reality. Classical approaches fall in between, with an $80$\% real-world success rate.

The takeaway for practitioners looking to build robots that navigate to objects is that the modular learning pipeline is very reliable, with a $90$\% success rate in limited time and efficient object search. In addition, we show that the remaining errors are primarily due to depth sensor failures (mapping failures and reflections in mirrors and TVs), which offers a clear path towards even greater reliability through better sensing or methods better able to deal with depth noise.

The takeaway for researchers is that much work remains to close the Sim-to-Real gap for semantic navigation. We identify two key issues that prevent today's simulators, 3D assets, and task definitions from being reliable evaluation benchmarks. Then we propose concrete steps along two orthogonal paths forward: improve sim to better reflect real-world conditions and improve practices to work with imperfect sim. 

First and foremost, there is a large Sim-to-Real gap in RGB images between sim and reality. Because of this, design choices easily overfit to sim. Two representative examples are that (A) policy architectures that directly operate on RGB images don't transfer because they overfit to sim images, and (B) the common practice of training segmentation models on sim data improves performance in sim but hurts in the real world. Improving sim to close this gap would involve increasing RGB photo-realism or providing extensive plug-and-play RGB randomization, both hard open problems~\cite{hofer2021sim2real}. This leaves us with no choice but to improve practices to work with today's sim. We should prioritize real-world transfer when designing policies: (A) replace policy architectures that directly operate on RGB images with ones leveraging abstractions as is common practice in other domains~\cite{muller2018driving, mahler2017dex, loquercio2021learning}, and (B) avoid training a segmentation model on sim data if the policy architecture does not allow easily swapping it for one trained on real-world data at inference time. 

Second, sim error modes don't accurately reflect real-world error modes, which limits the usefulness of sim to diagnose bottlenecks and further improve methods. Modular learning errors in the real world largely stem from depth sensor errors, while sim benchmarks usually assume perfect depth or don't provide realistic depth noise models. In contrast, errors in sim largely stem from reconstruction errors that do not happen in reality — both visual (imperfect RGB reconstruction that makes semantic segmentation harder in sim than reality) and physical (noisy navigation meshes that make planning harder in sim than reality). This explains the increase in performance from sim to reality for modular learning and stresses the need to always evaluate semantic navigation policies on real robots. We propose concrete steps forward to close this gap: (A) introduce realistic depth noise models for target deployment robots in sim benchmarks, (B) improve the visual quality of sim 3D scans, (C) improve the quality of sim navigation meshes. We hope our analysis sparks work to close the Sim-to-Real gap for semantic navigation.

\textbf{One-Sentence Summary:} Real-world empirical study of robot navigation methods comparing classical, end-to-end and modular learning approaches.

\section{Results} 

\href{https://www.youtube.com/watch?v=zZ6nlkTZVds&t=1s}{Movie 1} summarizes our results. We have deployed a policy representative of each of the classical, end-to-end learning, and modular learning approaches to Object Goal navigation on a Hello Robot Stretch robot~\cite{kemp2022design}. Stretch is a lightweight, compact, low-cost mobile manipulator with an RGB-D camera and LiDAR, which we use only for localization and collision avoidance. We evaluated the policies at scale over $60$ episodes split across six goal object categories in six different homes and a controlled study with one home replicated in sim. Before presenting our results, we give minimal formal background on the Object Goal task and the methods we evaluate.

\subsection{Navigating to Objects: Task and Approaches} 

\begin{figure}
    \centering
    \includegraphics[height=7.3cm,keepaspectratio]{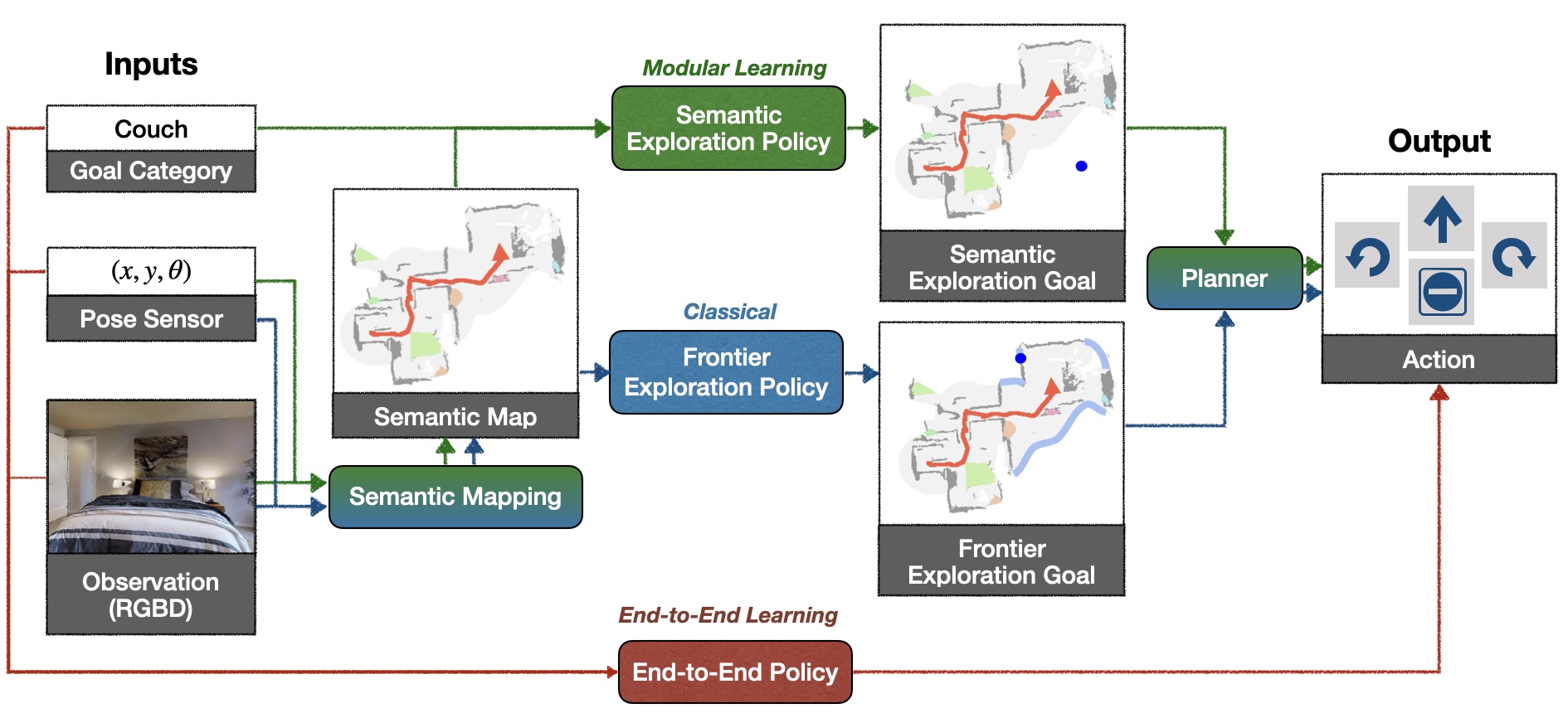}
    \caption{
    \small
    \textbf{Three approaches to navigate to objects.} \textbf{(A)} The modular learning approach builds a top-down semantic map, selects a semantic exploration goal in this space, and plans low-level actions to reach this goal.  \textbf{(B)} The classical approach also builds a semantic map but selects the closest unexplored region as the exploration goal, independently of the goal object.  \textbf{(C)} The end-to-end learning approach directly maps sensor inputs and the goal object to low-level actions with a deep neural network.}
    \label{fig:methods_overview}
\end{figure}

We consider the problem of semantic navigation instantiated by the Object Goal task~\cite{chaplot2020object, anderson2018evaluation, batra2020objectnav}. In the Object Goal task, the robot’s objective is to navigate to an instance of a particular object category (in our case, “chair”, “couch”, “potted plant”, “toilet”, “tv”, or “bed”) as efficiently as possible. The robot starts at a random location in an unknown home and receives the goal object category. At each step, the robot observes first-person RGB and depth images and takes a discrete navigation action: move forward ($25$ cm), turn left or right ($30$ degrees), or stop. The robot needs to take the stop action when it believes it has reached the goal object. An episode is considered successful if the robot’s distance to an instance of the goal object category is less than some threshold ($1$ meter) when the robot takes the stop action. In contrast, an episode is a failure if the agent calls the stop action too far from the goal object, never calls the stop action before a fixed maximum number of timesteps ($500$), or collides too many times (more than $20$) with its environment. We use two metrics for comparing methods: Success Rate (SR), the ratio of successful episodes, and Success weighted by Path Length (SPL)~\cite{anderson2018evaluation}, the ratio of path length over optimal path length for successful episodes, which measures exploration efficiency.
The Object Goal task requires spatial scene understanding (obstacle and navigable space detection), semantic scene understanding (object detection), learning semantic priors (for efficient exploration), and episodic memory (keeping track of explored and unexplored areas).

We evaluate three methods, each representative of one class of approaches. To represent modular learning for Object Goal navigation, we picked~\cite{chaplot2020object}. To represent classical approaches, we replace the semantic exploration policy of~\cite{chaplot2020object} with frontier exploration~\cite{yamauchi1997frontier}, which navigates towards the closest unexplored region. Finally, to represent end-to-end learning, we picked~\cite{ramrakhya2022habitat}. We will describe each method and what makes it representative of its class of approaches in detail in the Materials and Methods. For now, Fig.~\ref{fig:methods_overview} illustrates all the necessary background to understand the Results and Discussion. 

\subsection{Natural Home Environments} 

\begin{figure}
    \centering
    \includegraphics[height=6cm,keepaspectratio]{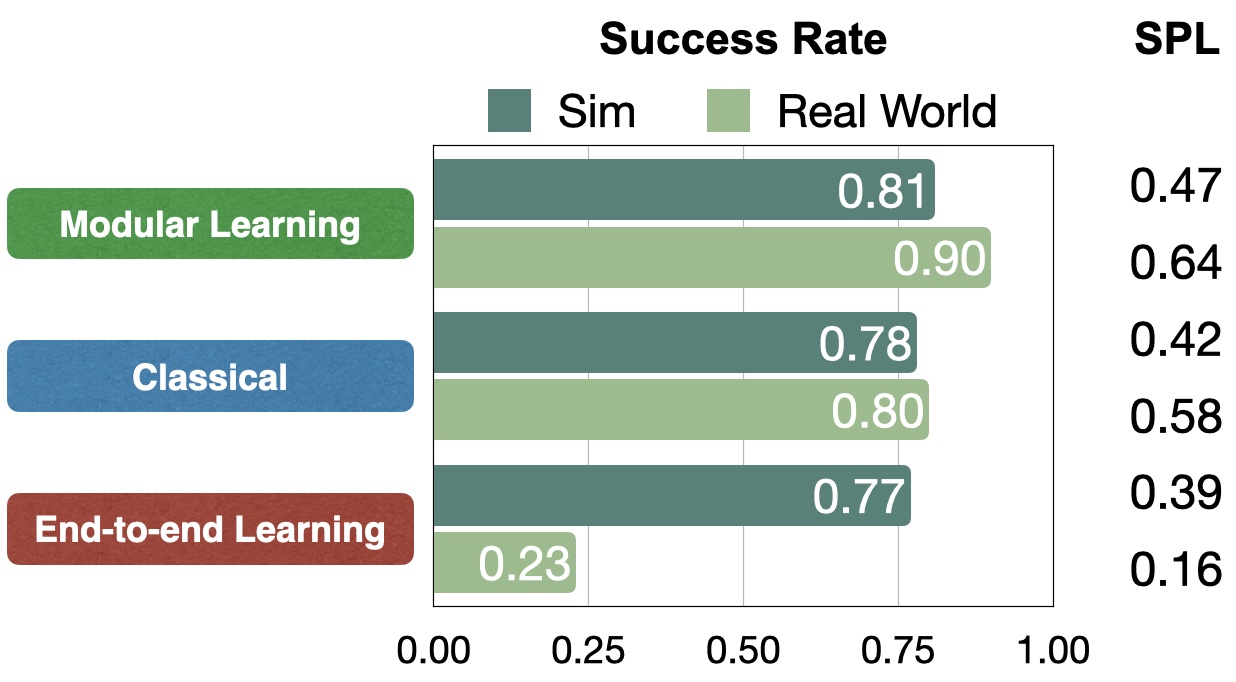}
    \caption{
    \small
    \textbf{Navigation performance in simulation vs. real at scale.} We compare the Success Rate (SR) and Success weighted by Path Length (SPL) of methods representative of classical, end-to-end-learning, and modular learning approaches on large real world ($60$ episodes in $6$ homes) and simulation datasets (single-floor navigation episodes of val split of the 2022 Habitat Challenge with $1093$ episodes in $20$ simulated homes of the HM3D Semantics dataset~\cite{yadav2022habitat}). \textbf{(A)} Performance for all methods is comparable in simulation, at around $80$\% success rate. \textbf{(B)} Classical and modular learning approaches transfer well, up from $78$\% to $80$\% and $81$\% to $90$\%, respectively. \textbf{(C)} End-to-end learning fails to transfer, down from $77$\% to $23$\% success rate.}
    \label{fig:main_results}
\end{figure}

We deployed navigation policies representative of each class of approaches in six natural home environments never seen before in simulation or reality. We evaluated the policies over $60$ episodes split across the six goal object categories and six homes. This represents $45$ hours of robot experiments ($3$ methods x $6$ homes x $10$ episodes per home x $15$ minutes per episode).

Fig.~\ref{fig:main_results} illustrates our main results quantitatively. Classical and modular learning approaches perform better in the real world than simulation, up from $78$\% to $80$\% and $81$\% to $90$\% success rate, respectively. In contrast, end-to-end learning performs much worse in the real world, down from $77$\% to $23$\% success rate. This trend holds across all homes and all goal object categories, as shown in Tables \ref{tab:real_world_results_details} and \ref{tab:real_world_results_aggregated}. Fig.~\ref{fig:all_methods1} and~\ref{fig:all_methods2} compare all approaches on the same episode to illustrate these results qualitatively. 
Fig.~\ref{fig:end_to_end_failures} illustrates two typical failure modes of the end-to-end learning policy, besides collisions. First, the policy often detects the goal object but fails to stop nearby, which is consistent with prior work observing this "last mile" failure in simulation~\cite{ye2021auxiliary}. Second, the policy revisits the same locations, often semantically unrelated to the goal object. These failure modes seem to display a lack of semantic understanding, a lack of long-term memory, and poor exploration.
Fig.~\ref{fig:learned_exploration_improvements} and \href{https://www.youtube.com/watch?v=zZ6nlkTZVds&t=1s}{Movie 1} illustrate how the modular learning approach improves over the classical approach. The learned exploration policy leverages the semantics of the top-down map to search for a specific goal object effectively. In contrast, the frontier exploration policy, which selects the closest unexplored region as the exploration goal independently of the goal object, exhibits depth-first search behavior and fails to backtrack using semantics.

\begin{figure}
    \thisfloatpagestyle{empty}
    \centering
    \includegraphics[height=19cm,keepaspectratio]{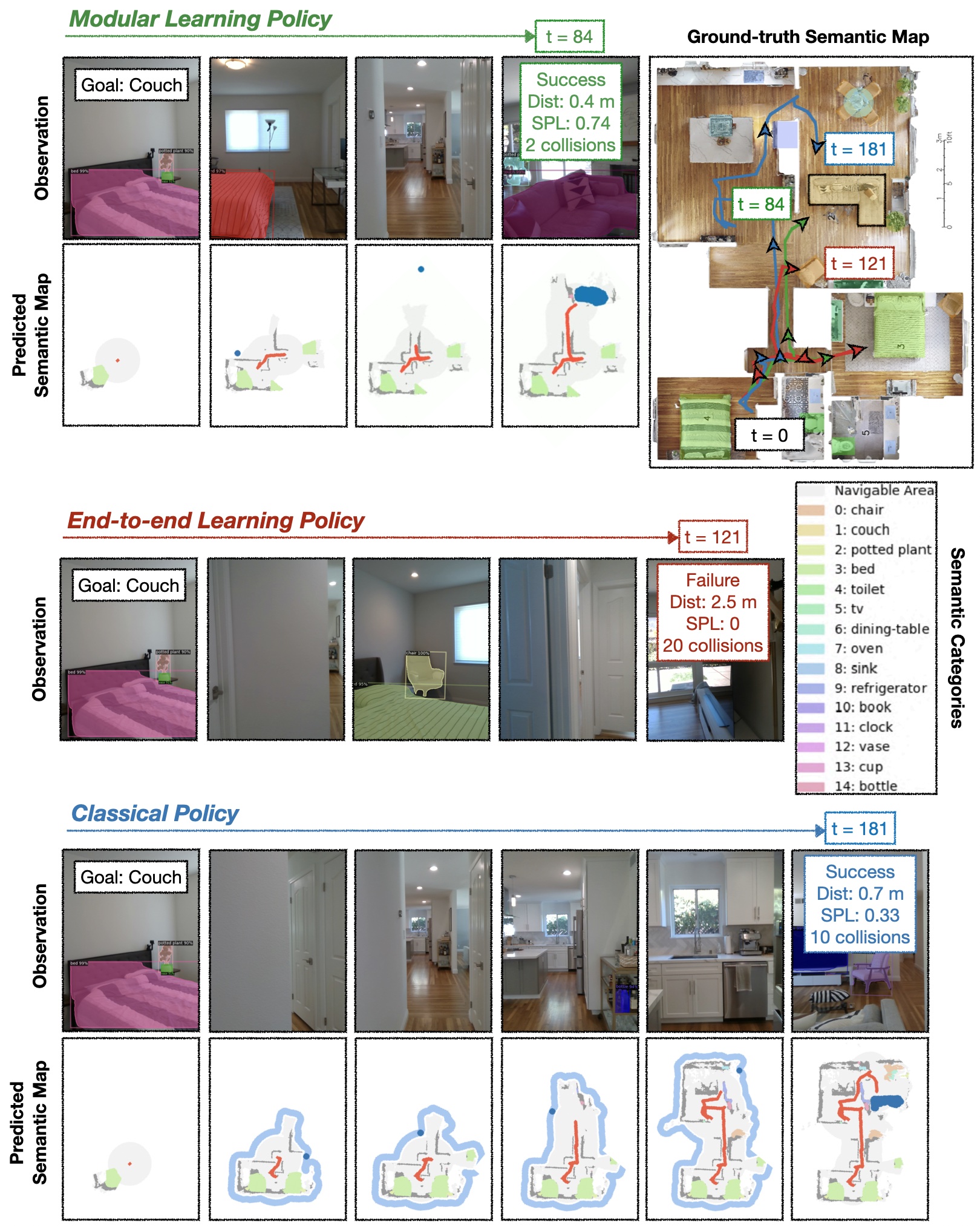}
    \caption{
    \small
    \textbf{Three approaches on the same episode.} \textbf{(A)} Modular learning reaches the couch goal in $84$ steps (SPL $= 0.74$). \textbf{(B)} End-to-end learning collides too many times ($20$ max) after $121$ steps. \textbf{(C)} The classical policy reaches the goal after $181$ steps and a detour through the kitchen (SPL $= 0.33$).}
    \label{fig:all_methods1}
\end{figure}

\subsection{Controlled Experiments in a Home Replicated in Simulation}

\begin{table*}[]
\caption{
    \small
	\textbf{Navigation performance in simulation vs. real on the same episodes in the same home.} 
        We compare policies by Success Rate (SR) on a sim benchmark (single-floor navigation episodes of val split of the 2022 Habitat Challenge with $1093$ episodes in $20$ simulated homes of the HM3D Semantics dataset~\cite{yadav2022habitat}), and the same home in sim and the real world ($10$ episodes). We compute the Sim-vs-Real Correlation Coefficient (SRCC) introduced in~\cite{habitatsim2real20ral}, which is the Pearson correlation between binary episode outcomes in sim vs. real (in $[0, 1]$, higher is better). \textbf{(A)} Top table: In an ablation study of end-to-end policies, we vary training camera parameters, the segmentation model training domain, and the policy training algorithm. Bottom table rows 1-4: Real-world performance of end-to-end policy ablations is inversely correlated to sim performance. Policy 1 is the best in sim but the worst in the real world, while policy 4 — which we selected for large-scale evaluation — is the worst in sim but the best in the real world. All variants other than policy 4 have a low SRCC, indicating their design overfits to sim. \textbf{(B)} Bottom table rows 5-7: Performance on the sim replica is close to that on the sim benchmark, which shows that our experimental setting matches that of the sim benchmark. Although, performance in the real-world home is close to that on its sim replica, the SRCC is still relatively low because policies fail different episodes in sim and the real world. As we will see in the Discussion, this is due to a disconnect between error modes in sim and reality.
}
\label{tab:controlled_study}
\centering
\small
\begin{tabular}{ccccc}
\multicolumn{5}{c}{\textbf{End-to-end learning ablations}} \\ \toprule\hline
\multicolumn{1}{c|}{\multirow{2}{*}{\textbf{\begin{tabular}[c]{@{}c@{}}End-to-end\\ Policy\end{tabular}}}} & \multicolumn{4}{c}{\textbf{Policy training settings}} \\ \cline{2-5} 
\multicolumn{1}{c|}{} & \textbf{Camera parameters} & \multicolumn{2}{c}{\textbf{Segmentation training domain}} & \textbf{Policy training algorithm} \\ \hline\midrule
\multicolumn{1}{l|}{\textbf{End-to-end 1}} & Sim Benchmark & \multicolumn{2}{c}{Sim} & IL + RL \\
\multicolumn{1}{l|}{\textbf{End-to-end 2}} & Robot & \multicolumn{2}{c}{Sim} & IL + RL \\
\multicolumn{1}{l|}{\textbf{End-to-end 3}} & Robot & \multicolumn{2}{c}{Real-world} & IL + RL \\
\multicolumn{1}{l|}{\textbf{End-to-end 4}} & Robot & \multicolumn{2}{c}{Real-world} & IL \\ \hline \bottomrule \\
\multicolumn{5}{c}{\textbf{Evaluation results}} \\ \toprule\hline
\multicolumn{1}{c|}{\textbf{\begin{tabular}[c]{@{}c@{}}Navigation \\ policy\end{tabular}}} & \textbf{Sim benchmark SR} & \textbf{\begin{tabular}[c]{@{}c@{}}Real home \\ sim replica SR\end{tabular}} & \textbf{\begin{tabular}[c]{@{}c@{}}Real home\\ SR\end{tabular}} & \textbf{\begin{tabular}[c]{@{}c@{}}Real home\\ SRCC\end{tabular}} \\ \hline \midrule
\multicolumn{1}{l|}{\textbf{End-to-end 1}} & 0.77 & 0.80 & 0.00 & 0.20 \\
\multicolumn{1}{l|}{\textbf{End-to-end 2}} & 0.71 & 0.70 & 0.00 & 0.30 \\
\multicolumn{1}{l|}{\textbf{End-to-end 3}} & 0.61 & 0.60 & 0.10 & 0.40 \\
\multicolumn{1}{l|}{\textbf{End-to-end 4}} & 0.48 & 0.50 & 0.30 & 0.60 \\ \hline
\multicolumn{1}{l|}{\textbf{\begin{tabular}[c]{@{}c@{}}End-to-end\end{tabular}}} & 0.48 & 0.50 & 0.30 & 0.60 \\
\multicolumn{1}{l|}{\textbf{\begin{tabular}[c]{@{}c@{}}Modular\end{tabular}}} & 0.81 & 0.80 & 0.90 & 0.70 \\
\multicolumn{1}{l|}{\textbf{Classical}} & 0.78 & 0.80 & 0.90 & 0.70 \\ \hline\bottomrule
\end{tabular}
\end{table*}

We ran a controlled study in one home replicated in simulation. This serves two purposes. First, this lets us decompose the discrepancy between results on a simulation benchmark and the real world into (A) the discrepancy between the sim benchmark and the sim replica and (B) the discrepancy between the sim replica and the real home. This lets us verify that our experimental setting matches the sim benchmark: results in our sim replica should be close to that of the sim benchmark. Second, this allows us to run an ablation study to select the best end-to-end policy to evaluate at scale across all six homes. We replicate a single home in simulation as 3D scanning and digitizing a house — with the same Matterport camera used to digitize homes of the sim benchmark~\cite{ramakrishnan2021habitat} — takes a few hours.

We first present the results of the ablation study we conducted to select the best end-to-end policy to evaluate at scale across all six homes. Modular learning and classical approaches only use first-person RGB-D and predicted segmentation images through a top-down semantic map. This makes them independent of changes in camera parameters and lets us easily swap semantic segmentation models between training and evaluation. In contrast, end-to-end learning approaches directly operate on RGB-D and predicted segmentation images. This means varying camera parameters between training and evaluation settings introduces a domain gap, and an end-to-end policy is closely tied to the segmentation model used in its network architecture during training. These characteristics of end-to-end policies mean that in order to give them the best chance to work on a robot, an ablation study is needed to select the best-performing policy.

Table \ref{tab:controlled_study} (A) shows results of this ablation study. We train four different end-to-end policies, varying the camera parameters, segmentation model training domain, and training algorithm. We measure each policy’s performance in sim, both on a large-scale benchmark and our sim replica, and in the real home. Overall, Policy 1, trained with the camera parameters of the sim benchmark, which we detail in the Materials and Methods section, and using a segmentation model trained in sim, performs the best in sim but the worst in the real world. Policy 2, trained with robot camera parameters, performs slightly worse in sim, which shows that our robot camera parameters make the task slightly harder than the sim benchmark camera parameters. Policies 3 and 4 that use a segmentation model trained in the real world perform worse in sim but better in the real world. This shows that using a segmentation model trained in sim is overfitting to sim. Policy 4, which is trained with imitation learning (IL) only as opposed to IL followed by reinforcement learning (RL) fine-tuning, performs the worst in sim but the best in reality. This shows that RL fine-tuning can overfit to sim. We select policy 4 to evaluate at scale across all six homes.

Overall, these results show that performance in sim is often inversely proportional to performance in the real world because design choices to improve performance in sim can easily overfit to sim. We can quantify this observation through the Sim-vs-Real Correlation Coefficient (SRCC) introduced in~\cite{habitatsim2real20ral}, which measures how well sim results can predict real-world results. The SRCC is low for all end-to-end policy variants, starting as low as $0.20$ for policy 1, which performs best in sim, and increasing to $0.60$ for policy 4 as we correct for design choices that overfit to sim. 

Now that we have selected the best end-to-end policy to evaluate at scale, Table \ref{tab:controlled_study} (B) decomposes the discrepancy between results for all selected methods on the sim benchmark and the real world into the discrepancy between the sim benchmark and the sim replica, and the discrepancy between the sim replica and the real home. First, performance on the sim replica is close to that on the sim benchmark, which verifies that our experimental setting matches that of the sim benchmark. Second, even though absolute performance is fairly close in sim and the real world for the policies we evaluate at scale (e.g., $0.80$ sim vs. $0.90$ real success rate for modular learning), the SRCC is still low (e.g., only $0.70$ for modular learning) because each method fails different episodes in sim and the real world. As we will see in the Discussion section, this is due to a disconnect between error modes in sim and reality.

\section{Discussion} 

In this section, we (A) analyze error modes of the modular learning approach in sim vs. real to make sense of its rise in performance from sim to real, (B) emphasize the significance of the $90$\% real-world success rate of modular learning, and (C) investigate why modular learning transfers better than end-to-end learning and reflect on what this means for the field.

\subsection{Modular Learning Error Modes in Real-world vs. Simulation}

To make sense of the rise from $81$\% sim to $90$\% real-world success rate for the modular learning approach, we compare its sim and real error modes in Table~\ref{tab:error_modes}. Surprisingly, Table~\ref{tab:error_modes} shows there is nearly no overlap between sim and real error modes. Errors in the real world largely stem from depth sensor errors ($5$ out of $6$ total errors), as illustrated in Fig.~\ref{fig:depth_sensor_errors} and \href{https://www.youtube.com/watch?v=zZ6nlkTZVds&t=1s}{Movie 1}, while the sim benchmark assumes perfect depth sensing. When approaching a door at an angle, noise in depth can block it in the map, making a room inaccessible without a map denoising mechanism. Reflection in mirrors and TVs can also cause depth sensor errors and downstream navigation failures. In contrast, a lot of the failures that occur in sim are due to reconstruction errors — both visual and physical — which do not happen in reality. Indeed, $10.1$\% out of the total $18.6$\% episode failures in the sim benchmark are due to segmentation errors, while segmentation errors did not cause any episode failures for modular learning in the real world. Segmentation errors are more common in sim because visual reconstruction can make objects unrecognizable, as illustrated in Fig.~\ref{fig:reconstruction_errors} (A) and \href{https://www.youtube.com/watch?v=zZ6nlkTZVds&t=1s}{Movie 1}. Physical reconstruction errors represent another $5.5$\% of the total $18.6$\% episode failures in sim. They lead to noisy navigation meshes with narrow pathways that are hard to navigate for discrete planners that work well in the real world, as illustrated in Fig.~\ref{fig:reconstruction_errors} (B) and \href{https://www.youtube.com/watch?v=zZ6nlkTZVds&t=1s}{Movie 1}. This gap in error modes explains the performance gap between sim and reality for modular learning. 

The lack of overlap between sim and real-world error modes is concerning because it limits the usefulness of simulation to diagnose bottlenecks and further improve policies. This is a practical working definition of the Sim-to-Real gap: we only care about improving sim realism to the extent that it lets us develop better real-world policies. A gap in error modes prevents us from doing so. This stresses the need to always evaluate semantic navigation policies on real robots for results to be meaningful. Based on our analysis, we propose concrete steps forward to close this gap: (A) introduce realistic depth noise models for target deployment robots in sim benchmarks, (B) improve the visual quality of sim 3D scans, (C) improve the quality of sim navigation meshes.

\subsection{Towards Solving Navigation to Objects with Modular Learning}

The main takeaway of this study for practitioners looking to build robots that navigate to objects is that the modular learning pipeline is very reliable, with a $90$\% success rate in limited time and efficient object search with an SPL of $0.64$. In addition, we show that the remaining errors are primarily due to depth sensor failures, which offers a clear path towards even greater reliability through better sensing or methods better able to deal with depth noise.

One limitation of our study is the restriction to six goal object categories to align with current sim benchmarks. All methods we evaluate are straightforward to extend to a larger finite set of categories, and opportunities for future work include extensions to an unbounded set of object categories with open-vocabulary detectors~\cite{zhou2022detecting} and instance-level object goals~\cite{li2021ion}.

\subsection{Why Modular Learning Transfers Better than End-to-end Learning}

\begin{figure}
    \centering
    \includegraphics[height=10cm,keepaspectratio]{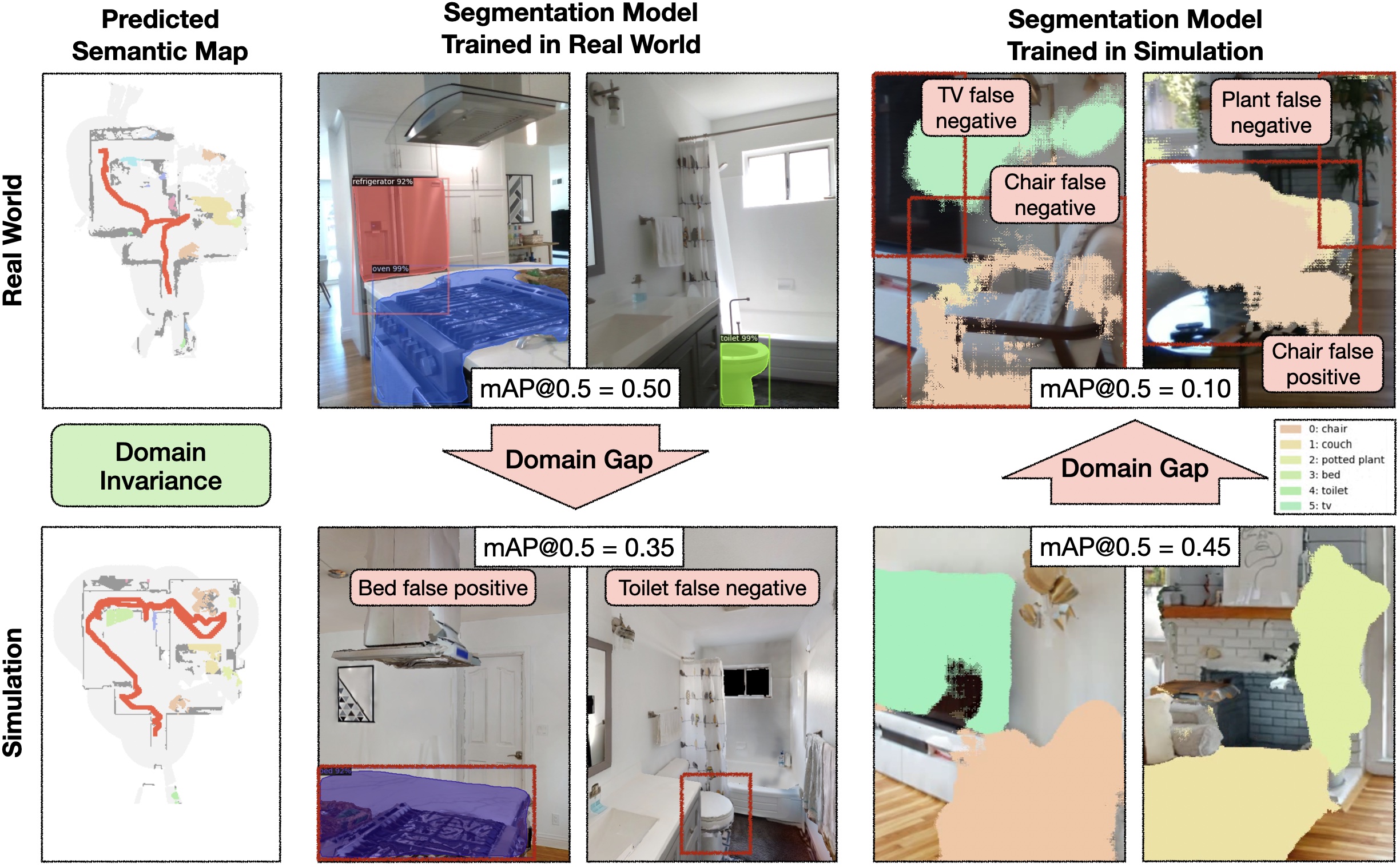}
    \caption{
    \small
    \textbf{Sim-vs-Real domain invariances, gaps, and their effects on segmentation.} From left to right, all images come from episodes in our controlled study: \textbf{(A)} The semantic map space is invariant between the real world and simulation. \textbf{(B)} The image space exhibits a large gap between the real world and simulation. \textbf{(C)} This gap causes a large drop in performance when transferring a segmentation model trained in the real world to simulation and vice versa.}
    \label{fig:domain_gap}
\end{figure}

Fig.~\ref{fig:domain_gap} illustrates why modular learning transfers better than end-to-end learning. The semantic exploration policy of the modular learning approach takes a semantic map as input, while the end-to-end policy directly takes the RGB-D frames as input. The figure shows that the semantic map space is invariant between simulation and the real world. This is due to the semantic map abstracting away pixels in favor of semantic categories and reducing the spatial granularity of the map voxel size ($5$ cm in our experiments). In contrast, the RGB image space exhibits a large gap between the real world and simulation because current reconstruction engines cannot yet generate photo-realistic images. This causes a large drift between the training and evaluation domains for the deep neural network of the end-to-end policy. 

The significance of the gap from simulation images to real-world images is well illustrated by the segmentation errors it causes in Fig.~\ref{fig:domain_gap}. A segmentation model trained in the real world suffers a severe performance drop when transferred to simulation, down from $0.50$ to $0.35$ mean average precision at a $0.50$ intersection over union threshold (mAP@$0.5$). Similarly, a segmentation model trained in simulation suffers an even larger performance drop when transferred to the real world (down from $0.45$ to $0.10$ mAP@$0.5$). If semantic segmentation transfers poorly from simulation to reality, it is reasonable to expect an end-to-end semantic navigation policy trained on sim images to transfer poorly to real-world images. While the image domain gap affects a segmentation model over a single prediction, the gap accumulates over many predictions made over a long horizon for an end-to-end navigation policy.

Because of this image domain gap, design choices easily overfit to simulation. Two representative examples in our results are that (A) end-to-end policy architectures that directly operate on RGB images don't transfer because they overfit to simulation images, and (B) the common practice of training segmentation models on simulation data improves performance in simulation but hurts in the real world.

How can we address this image domain gap? At this point, it is worth taking a step back to briefly survey how other domains of robotics tackle Sim-to-Real gaps. Prevalent options today~\cite{hofer2021sim2real} are: (A) train on real-world data, (B) train in simulation with domain randomization, and (C) train in simulation with modularity and abstraction. Training on real-world data bypasses any Sim-to-Real gap but is expensive, slow, and potentially unsafe. It is the option of choice for perception stacks across robotics and has been applied to end-to-end control in domains where sub-optimal policy operation is safe, like grasping and static manipulation~\cite{pinto2016supersizing, kalashnikov2018scalable}. But it is not straightforward to scale up for semantic navigation as mobile robots require constant supervision. This leaves us with simulation training. Domain randomization ~\cite{tobin2017domain} — randomizing environmental factors that are hard to simulate accurately during training to make policies robust to these factors — has been applied successfully in domains where randomization is straightforward like dexterous manipulation~\cite{andrychowicz2020learning} where one needs to randomize over physics and single object appearance. But it is still an open problem whether RGB image randomization can be scaled up to entire houses to bring the same robustness to semantic navigation~\cite{hofer2021sim2real}. Our final option, training in simulation with modularity and abstraction~\cite{muller2018driving} — designing abstractions of the input raw sensor data that contain sufficient information to solve the task while being invariant to environmental factors that are hard to simulate accurately — is the practice of choice in autonomous flight~\cite{kaufmann2020deep, loquercio2021learning}, legged locomotion~\cite{miki2022learning, fu2022coupling}, grasping~\cite{mahler2017dex, mousavian20196}, and a promising path in autonomous driving~\cite{muller2018driving}.

In our view, training in simulation and Sim-to-Real transfer via modularity and abstraction is the most promising path forward for semantic navigation. Example semantic abstractions could be first-person semantic segmentation masks~\cite{muller2018driving}, topological scene representations~\cite{savinov2018semi}, or top-down spatial semantic maps~\cite{chaplot2020object}. One limitation of our study and opportunity for future work is that we didn't take further steps in this direction — like evaluating an end-to-end policy abstracting RGB frames through semantic frames.

\section{Materials and Methods} 

\begin{figure}
    \thisfloatpagestyle{empty}
    \centering
    \includegraphics[height=15cm,keepaspectratio]{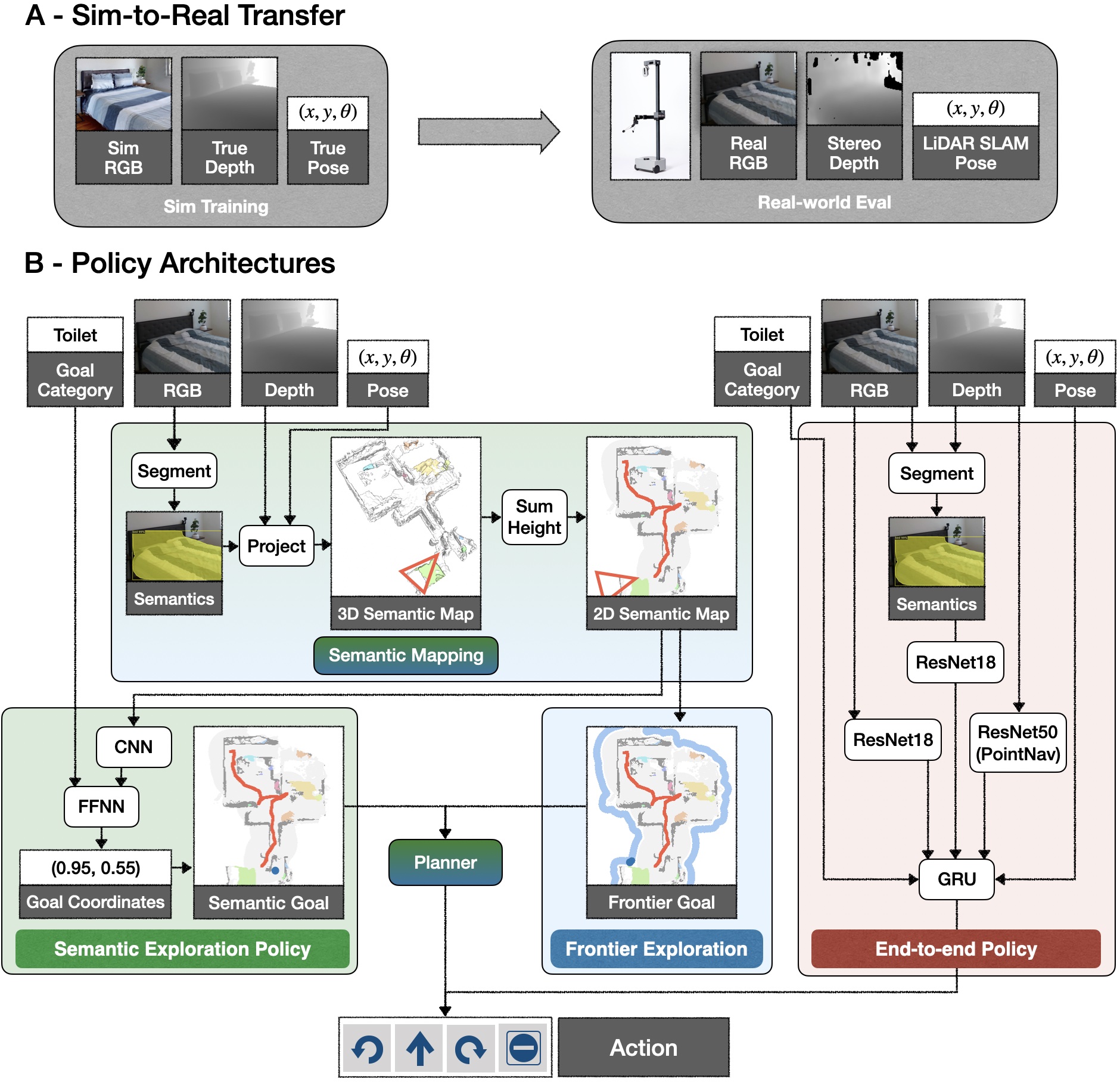}
    \caption{
    \small
    \textbf{Three approaches to navigate to objects and their Sim-to-Real transfer.} \textbf{(A)} We match policy inputs from sim in the real world: we estimate depth with a stereo camera, and the pose with LiDAR-based SLAM. \textbf{(B)} Policy architectures in detail from left to right, top to bottom. The semantic mapping module of classical and modular learning approaches segments the RGB frame, projects it into a 3D semantic voxel map using the depth and pose, and sums over the height to compute a 2D semantic map. The semantic exploration policy of the modular learning approach predicts an exploration goal with a feed-forward neural network (FFNN) as a function of map features computed with a convolutional neural network (CNN) and the goal object. The frontier exploration policy of the classical approach explores the closest unexplored region independently of the goal object. Both approaches plan low-level actions to reach the high-level exploration goal. The end-to-end approach segments the RGB frame with a pre-trained segmentation model, computes RGB, depth, and semantic features with CNN backbones, which can be also be pre-trained, and directly predicts low-level actions from frame features, the goal object, and the pose with a Gated Recurrent Unit (GRU).}
    \label{fig:methods_details}
\end{figure}

We deployed three navigation policies on a robot across six homes to study how well different methods to navigate to object categories transfer from simulation to the real world. This section outlines our Sim-to-Real transfer methodology, details the specific methods we evaluate, and explains what makes them representative of broader classes of approaches.

\subsection{Sim-to-Real Transfer Methodology}

\noindent\textbf{Hardware and Software Stack:} We deployed navigation policies on a Hello Robot Stretch robot~\cite{kemp2022design}. We selected Stretch because it is low-cost, lightweight, and compact, allowing us to transport it out of the lab into real homes easily. We trained all learned components of navigation policies in simulation with the Habitat platform~\cite{szot2021habitat}. We selected Habitat for its simulation speed, which is crucial for training navigation policy components with reinforcement learning. At inference time, policies were deployed with the fairo library~\cite{fairo2021} to run the same navigation code in  simulation and the real world.

\vspace{2mm}
\noindent\textbf{Matching Policy Inputs and Outputs from Sim to Real:} All navigation policies we evaluated were trained in simulation and had so far only been evaluated in simulation. To transfer these policies to a robot and be able to compare real-world results to simulation, we had to match the Object Goal task input and output spaces from sim on the robot. In the Object Goal task, the robot receives an RGB-D image and a sensor pose at each step. Stretch’s Intel RealSense D435i camera gives us ($640$ x $480$) RGB-D images with a 42-degree horizontal field of view. We preprocess the depth with standard spatial and temporal filters and threshold all values beyond the camera's confidence range (4 meters). We use off-the-shelf LiDAR-based Hector SLAM~\cite{KohlbrecherMeyerStrykKlingaufFlexibleSlamSystem2011} to estimate the robot's pose. We also use LiDAR for collision detection to deploy policies safely. To guarantee a fair comparison to simulation settings, we restrict the use of LiDAR to collision detection and localization and do not use it for mapping. Given the output action space is discrete — move forward $25$ cm, turn left/right $30$ degrees, or stop — it is straightforward to match on the robot. Sim-to-Real transfer is illustrated in Fig.~\ref{fig:methods_details} (A).

\vspace{2mm}
\noindent\textbf{Generating Episodes in Real Homes:} We evaluate the robot on $60$ episodes across six object goal categories (“chair”, “couch”, “potted plant”, “toilet”, “tv”, and “bed”) in six homes. We selected these goal object categories to be able to directly compare results in the real world with that of a leading sim benchmark: the Habitat 2022 Object Navigation Challenge. We selected homes for their visual diversity and size. To generate an episode within a home, we selected a goal category and a starting location for the robot while trying to balance the distribution of goal object categories and match the distribution of geodesic distances to the goal to that of the Habitat Challenge, as shown in  Fig.~\ref{fig:trajectory_statistics}. We consider an episode successful if the robot called the stop action close enough (less than 1 meter) to an instance of the goal category within the time ($200$ steps) and collision ($20$ collisions) budgets. To compute the Success weighted by Path Length (SPL)~\cite{anderson2018evaluation}, we measure the geodesic distance to the goal object instance closest to the starting location. To replicate one home and its episodes in simulation in our controlled study, we scanned and digitized the home with a Matterport Pro2 3D camera and matched the real-world goals and starting positions in sim.

\vspace{2mm}
Fig.~\ref{fig:methods_details} (B) illustrates the architectures of the three policies we evaluate. We first describe the classical and modular learning semantic mapping approaches before turning our attention to end-to-end learning approaches.

\subsection{Classical and Modular Learning Approaches}

\noindent\textbf{Extending Classical Approaches to Navigate to Objects:}
Classical simultaneous localization and mapping (SLAM) approaches to spatial navigation typically build a spatial map of the environment while simultaneously localizing the robot relative to its growing map~\cite{thrun2001robust, thrun2002robotic}, and navigate to geometric points within this map via path planning. Adapting these methods to navigate to objects requires detecting objects, keeping objects in memory, and exploring semantically towards objects. Semantic SLAM methods~\cite{flint2011manhattan, kundu2014joint, bowman2017probabilistic, ma2017multi, zhang2018semantic, rosinol2020kimera} naturally extend SLAM to detect objects and keep them in memory in a spatial semantic map but offer no solution for efficient semantic exploration. Among the many heuristics for goal-agnostic exploration proposed in the classical navigation literature, we select frontier-based exploration~\cite{yamauchi1997frontier} — navigate towards the closest unexplored region — to represent classical exploration methods because it was particularly effective in prior work~\cite{chaplot2020object}. As shown in Fig.~\ref{fig:methods_details} (B), this completes a pipeline representative of classical approaches: semantic mapping to keep seen objects in memory, frontier exploration to select high-level goals, and path planning to select low-level actions. 

Note that if we only care about navigating to a single object starting with zero information about the environment — the setting we evaluate in this paper — we don’t even need to keep objects in memory in a semantic map in the classical approach. We can simply use a geometric map for exploration and planning and head toward the goal object as soon as we detect it in the frame with a pre-trained object detector. We add semantic mapping to the classical approach for ease of implementation — we can head towards an object by projecting it into a single semantic channel in the map and planning toward it — and to make it applicable to the more practical but harder-to-evaluate setting where the robot might need to execute several such object search commands in a row.

\vspace{2mm}
\noindent\textbf{Modular Learning to Navigate to Objects:} 
While frontier exploration is effective, as shown in our results, it is necessarily suboptimal because it ignores the goal object. We intuitively understand where a “couch” is more likely to be found. This is where modular learning comes into play: can we train an exploration policy to leverage the statistical regularities in the layout of objects in homes to explore more efficiently? This is what the semantic exploration method proposed in~\cite{chaplot2020object} does. We selected it to represent modular learning as it cleanly isolates the problem of learning an exploration module from the rest of the navigation problem. As shown in Fig.~\ref{fig:methods_details} (A), it preserves the structure of the classical pipeline to navigate to objects presented above but replaces frontier exploration with a learned semantic exploration module. With this high-level picture in mind, we explain each component in detail.

\vspace{2mm}
\noindent\textbf{Semantic Map Representation:} The semantic map is a spatial representation of the environment that keeps track of objects, obstacles, and explored areas. Concretely, it is a binary $K$ x $M$ x $M$ matrix where $M$ x $M$ is the map size and $K$ is the number of map channels. Each cell of this spatial map corresponds to $25$ cm$^2$ ($5$ cm x $5$ cm) in the physical world. Map channels $K = C + 4$ where $C$ is the number of semantic object categories, and the remaining $4$ channels represent the obstacles, the explored area, and the agent’s current and past locations. An entry in the map is one if the cell contains an object of a particular semantic category, an obstacle, or is explored, depending on the channel, and zero otherwise. The map is initialized with all zeros at the beginning of an episode and the agent starts at the center of the map facing east.

\vspace{2mm}
\noindent\textbf{Semantic Mapping Module:} In order to build the semantic map, we need to predict semantic categories and segmentation masks of objects in first-person observations. We use a Mask-RCNN~\cite{mask_rcnn} with ResNet50~\cite{he2016deep} backbone pretrained on MS-COCO for object detection and instance segmentation. We project first-persons semantic segmentation into a point cloud using the depth, bin the point cloud into a 3D semantic voxel map, transform it from the robot’s coordinate system to that of the semantic map using the robot pose, and finally sum over the height to compute a 2D semantic map.

\vspace{2mm}
\noindent\textbf{Frontier Exploration Policy:} With our semantic map at hand, it is straightforward to implement frontier exploration: each step, we select the boundary between the explored and unexplored region of the map, i.e., the frontier, and within this boundary select the point closest to the robot in geodesic distance. This exploration strategy exhibits depth-first search behavior: once the robot heads into a direction, the closest unexplored region is in front of it until an obstacle blocks the way. As soon as the goal object is seen, i.e., as soon as the channel of the semantic map for the object goal has a nonzero element, we stop exploring and select all non-zero elements as the goal.

\vspace{2mm}
\noindent\textbf{Semantic Exploration Policy:} The semantic exploration strategy decides a goal as a function of the current semantic map and the goal object. This requires learning semantic priors on the relative arrangement of objects and areas in homes. As shown in Fig.~\ref{fig:methods_details} (B), map features are computed with a convolutional neural network (CNN) and passed through a feed-forward neural network (FFNN) along with a learnable embedding for the goal object to compute a goal in $[0, 1]^2$ , which is then converted to top-down map space. The policy is trained using reinforcement learning (RL) with the distance reduced to the nearest goal object as the reward. We use a policy trained in~\cite{chaplot2020object} on home layouts from the Gibson dataset~\cite{xia2018gibson}. As in~\cite{chaplot2020object}, we sample the long-term goal at a coarse time-scale, once every $25$ steps. This reduces the time-horizon for exploration in RL exponentially and consequently, reduces the sample complexity. As for the frontier exploration policy, as soon as the goal object is seen, we stop exploring and select it as the goal.

\vspace{2mm}
\noindent\textbf{Planner:} Given a long-term goal output by the frontier or semantic exploration policy, we use the Fast Marching Method~\cite{sethian1996fast}  as in~\cite{chaplot2020object} to plan a path and the first low-level action along this path deterministically. Although the semantic exploration policy acts at a coarse time scale, the planner acts at a fine time scale: every step we update the map and replan the path to the long-term goal.

\vspace{2mm}
\noindent\textbf{Sim-to-Real Transfer:} Semantic mapping methods are straightforward to transfer to the robot as they are independent of camera parameters and the semantic map space is invariant between sim and the real-world, as presented in the Discussion section.

\subsection{End-to-end Learning Approaches}

\noindent\textbf{End-to-end Learning to Navigate to Objects:} 
In contrast to modular approaches that explicitly build a semantic map of the environment and plan actions, end-to-end learning approaches directly predict actions from raw sensor data with a deep neural network. The network needs to learn to understand the scene spatially and semantically, keep track of long-term memory, and plan actions. Given each of these tasks is hard in isolation, end-to-end learning approaches typically require tens of thousands of expert demonstrations or hundreds of millions of steps of reinforcement learning to train. Given neither of these can realistically be collected in the real world for our Object Goal navigation task, training must be done in simulation. As illustrated in Fig.~\ref{fig:methods_details} (B), at each step, the typical end-to-end architecture for long horizon tasks with high-dimensional image input computes image features with one or multiple convolutional neural networks, which can be trained from scratch or pre-trained, and keeps track of memory and plans implicitly with a recurrent neural network, in our case a gated recurrent unit (GRU)~\cite{chung2014empirical}. 

\vspace{2mm}
\noindent\textbf{Habitat-Web Policy:} 
We selected Habitat-Web~\cite{ramrakhya2022habitat} as a representative example of end-to-end learning for the Object Goal navigation task because it closely follows the above paradigm and has the highest performance on a leading sim benchmark, the 2022 Habitat Challenge, at the time of publication. Habitat-Web~\cite{ramrakhya2022habitat} trains an end-to-end policy from human demonstrations on the Object Goal task in the Habitat simulator. We preserve the general architecture of the policy and swap different components in our ablation study to find the policy that works best in the real world. Semantic segmentation is predicted from RGB (and possibly depth) with a pre-trained and frozen segmentation model. First-person RGB, depth, and semantic frame features are computed with multiple CNN backbones, leveraging pre-trained models when available to reduce sample complexity. RGB, depth, and semantic frames are then fed into a GRU~\cite{chung2014empirical} along with other low dimensional features — the robot pose, a learnable goal object embedding, the fraction of the visual input occupied by the goal category, and the previous action — to predict a distribution over next actions. 

\vspace{2mm}
\noindent\textbf{Architecture and Training Details:} 
The segmentation model used in the original architecture is a RedNet~\cite{jiang2018rednet} pre-trained on the SUN RGB-D dataset~\cite{song2015sun} and fine-tuned on images from the Habitat simulator. As shown in our results, using a segmentation model fine-tuned in simulation hurts performance in the real world. In our ablation study of end-to-end architectures, we thus replace this model with the same Mask-RCNN pre-trained on MS-COCO used by modular semantic mapping methods. Depth features are computed with a ResNet50 pre-trained on Point Goal navigation~\cite{wijmans2019dd}. RGB and semantic frame features are computed with a ResNet18 trained from scratch.  The network is first trained with imitation learning (IL) on 80,000 human demonstrations (or approximately 20 million actions) collected for the Object Goal task in the Habitat simulator over three days with 128 Nvidia V100 32GB GPUs. It is then fine-tuned with reinforcement learning (RL) with a sparse binary reward when the goal is found, over 150 million steps in an additional three days on 32 Nvidia V100 32GB GPUs. Our ablation study of end-to-end policies shows that RL fine-tuning helps in sim but hurts in the real world.

\vspace{2mm}
\noindent\textbf{Sim-to-Real Transfer:} 
While semantic mapping methods are independent of camera parameters, end-to-end methods directly operate on first-person frames, and any change in camera parameters causes a large domain gap. Given our robot camera parameters are different from those used to train the original Habitat-Web policy — ($640$ x $480$) frames with a $42 \degree$ horizontal field of view, as opposed to ($480$ x $640$) frames with a $79 \degree$ horizontal field of view — we retrain the policy with robot camera parameters replicated in simulation.  

As shown in Fig.~\ref{fig:methods_details} (A), real-world depth estimated via stereo camera is noisy, which introduces an additional domain gap. To compensate for this, we tried training with the indoor depth noise model provided by Habitat~\cite{choi2015robust}, as we didn’t have any noise model tuned specifically for our robot’s Intel RealSense D435i camera. We found this to hurt real-world performance, suggesting this noise model doesn’t match our robot’s camera noise. 

While we can easily swap a different segmentation model in modular semantic mapping methods, changing the segmentation model causes a large domain gap for end-to-end methods and requires training a new policy. 

In contrast to simulation, robot discrete actions are also not deterministic in the real world due to actuation noise: a $25$ cm forward move or $30$ degree turn command will not have the exact intended effect. The usual way to compensate for this is to simulate actuation noise during training. We did not simulate actuation noise because a prior Sim-to-Real study of Point Goal navigation~\cite{habitatsim2real20ral} found this to hurt real-world performance. 

Finally, while we can easily inspect the output of various modules — like the semantic map, the exploration goals, or the plans — to diagnose issues when transferring modular methods, our only recourse when transferring end-to-end policies is to try matching training and evaluation inputs as closely as possible. All these considerations make transferring end-to-end policies much more challenging than modular methods.

\section{Acknowledgments} We thank Saurabh Gupta for reviewing the manuscript and Brandon Trabucco for lending his voice to the video.


\section{Supplementary Materials}
\beginsupplement

Fig.~\ref{fig:learned_exploration_improvements}. Comparison of modular learning and classical policies.\\
Fig.~\ref{fig:all_methods2}. Three approaches on the same episode.\\
Fig.~\ref{fig:end_to_end_failures}. Failure modes of end-to-end learning.\\
Fig.~\ref{fig:depth_sensor_errors}. Real-world depth sensor error modes.\\
Fig.~\ref{fig:reconstruction_errors}. Simulation visual and physical reconstruction error modes.\\
Fig.~\ref{fig:trajectory_statistics}. Distributions of goal objects and geodesic distances to goal in reality vs. sim.\\ 
Table~\ref{tab:real_world_results_details}. Navigation performance on all 60 episodes.\\
Table~\ref{tab:real_world_results_aggregated}. Navigation performance aggregated by home and goal object.\\
Table~\ref{tab:error_modes}. Modular learning real vs. sim error modes.\\
Table~\ref{tab:programmatic_error_analysis}. Modular learning programmatic error analysis on sim benchmark.\\
Table~\ref{tab:manual_error_analysis}. Modular learning manual analysis of remaining errors on sim benchmark.\\
Table~\ref{tab:error_analysis_episode_outcomes}. Modular learning episode outcomes on sim benchmark.\\

\begin{figure}[h]
    \centering
    \includegraphics[height=12cm,keepaspectratio]{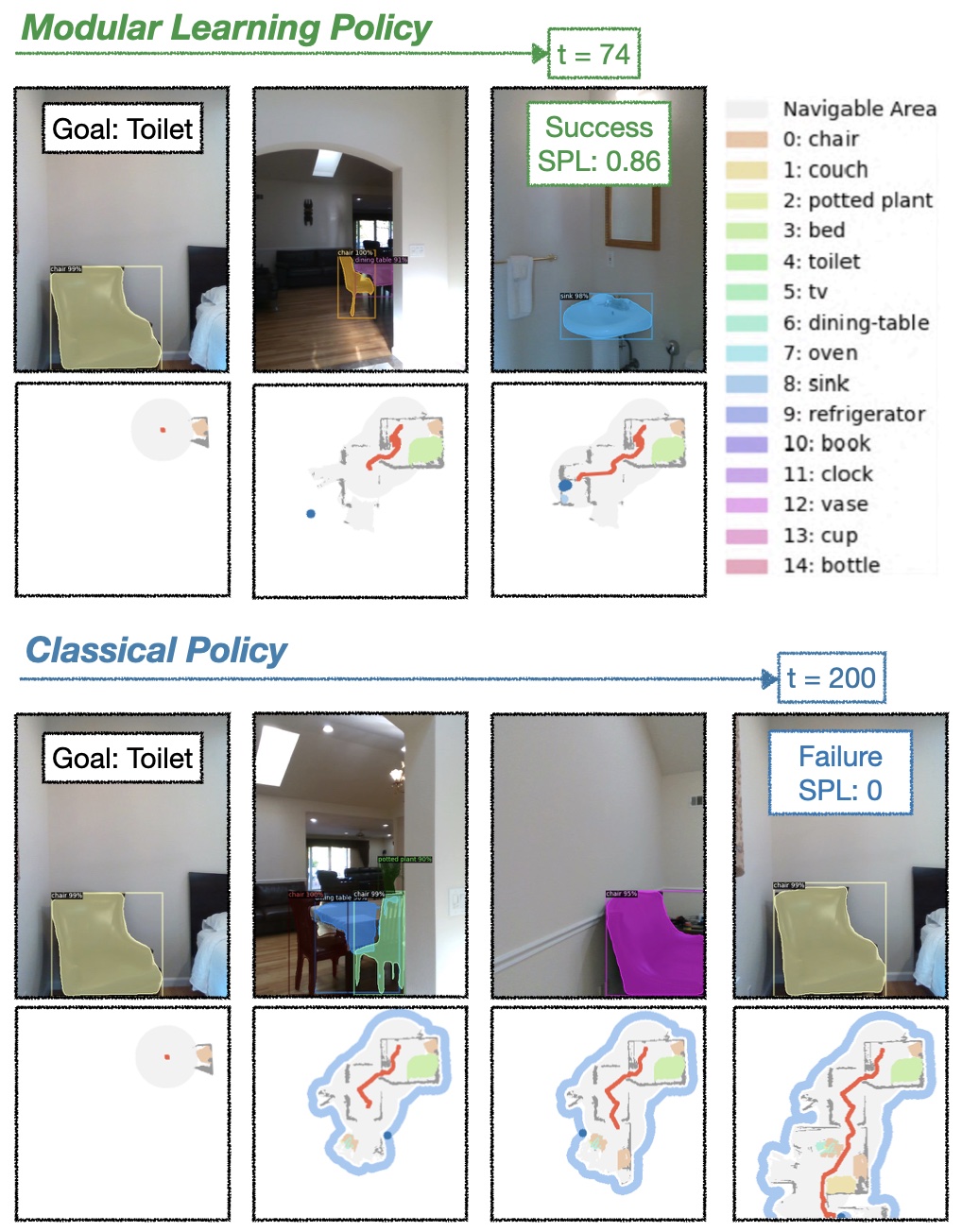}
    \caption{\textbf{Comparison of modular learning and classical policies.} \textbf{(A)} Top: the learned semantic exploration of the modular learning approach finds the toilet goal in only $74$ steps. \textbf{(B)} Bottom: the frontier exploration of the classical approach fails to find the toilet goal in $200$ steps.}
    \label{fig:learned_exploration_improvements}
\end{figure}

\begin{figure}
    \thisfloatpagestyle{empty}
    \centering
    \includegraphics[height=19cm,keepaspectratio]{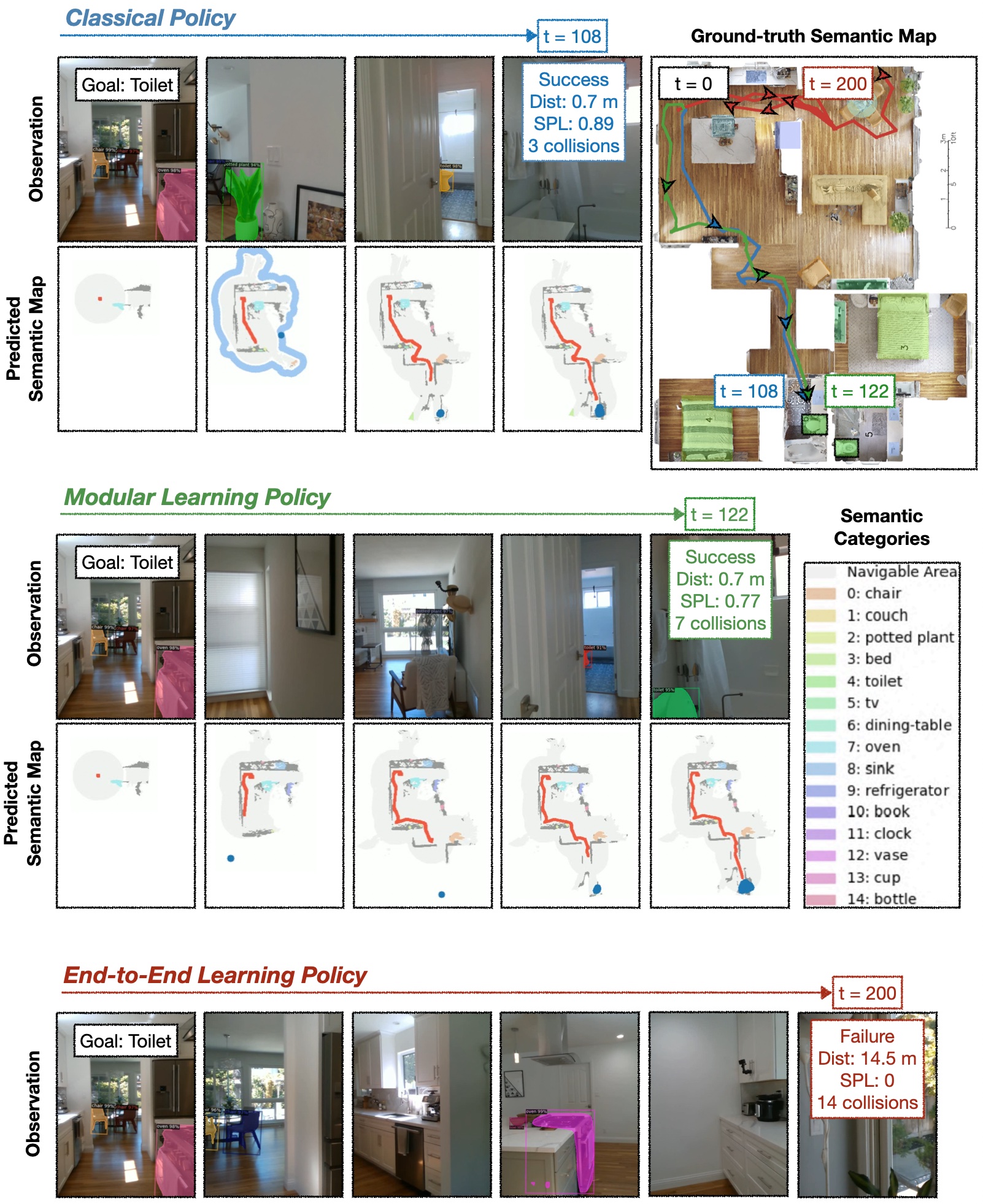}
    \caption{
    \small
    \textbf{Three approaches on the same episode.} \textbf{(A)} Both the classical and modular learning policies reach the toilet goal efficiently, in $108$ and $122$ steps, respectively. \textbf{(B)} The end-to-end learning policy fails to explore beyond the kitchen and dining room and never reaches the toilet goal.}
    \label{fig:all_methods2}
\end{figure}

\begin{figure}[h]
    \centering
    \includegraphics[height=11cm,keepaspectratio]{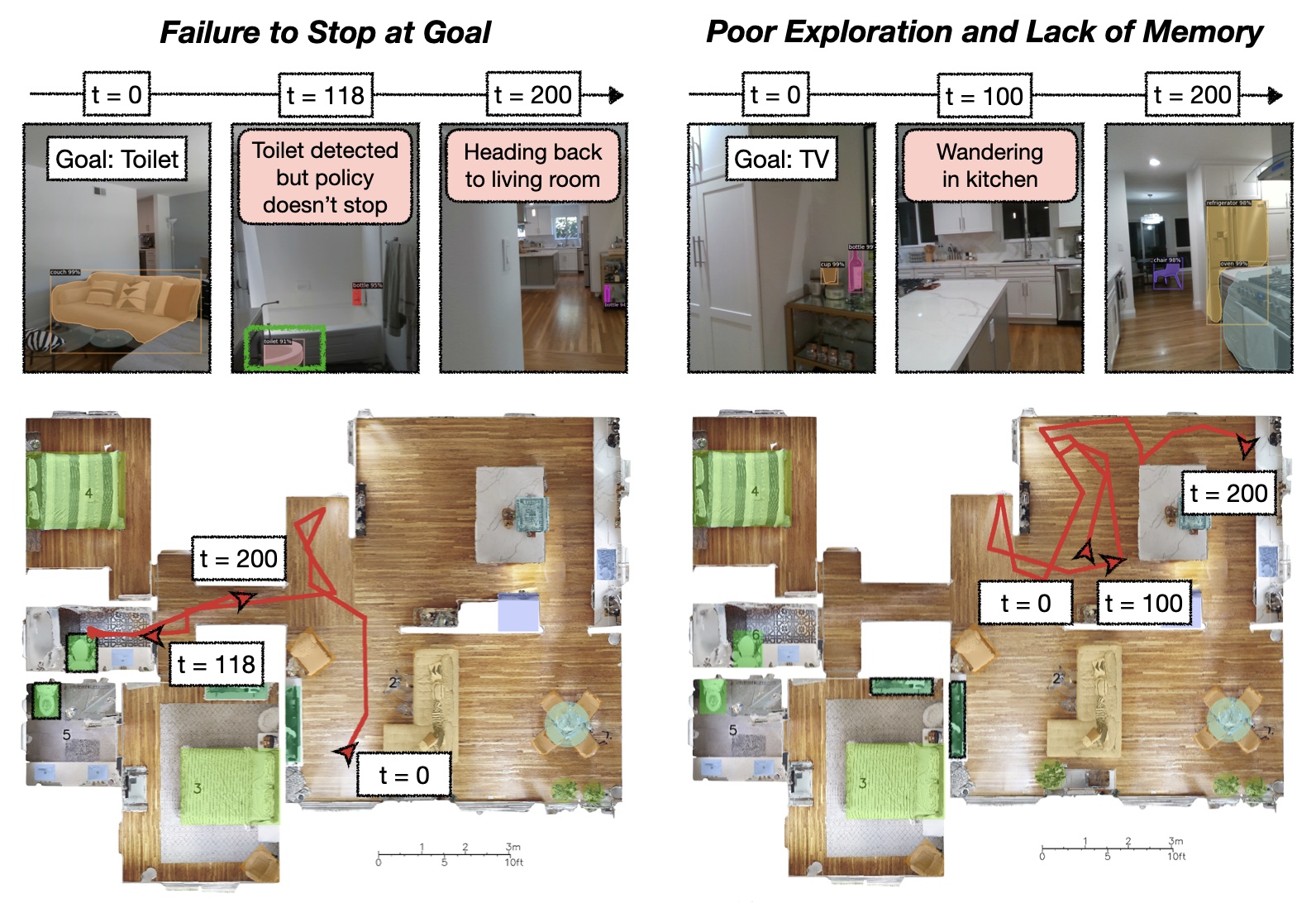}
    \caption{
    \textbf{Failure modes of end-to-end learning.} \textbf{(A)} The end-to-end policy detects the toilet goal but fails to stop and heads back towards the living room. \textbf{(B)} The end-to-end policy fails to explore effectively, revisiting the same locations in the kitchen while looking for a TV.}
    \label{fig:end_to_end_failures}
\end{figure}

\begin{figure}[h]
    \thisfloatpagestyle{empty}
    \centering
    \includegraphics[height=20cm,keepaspectratio]{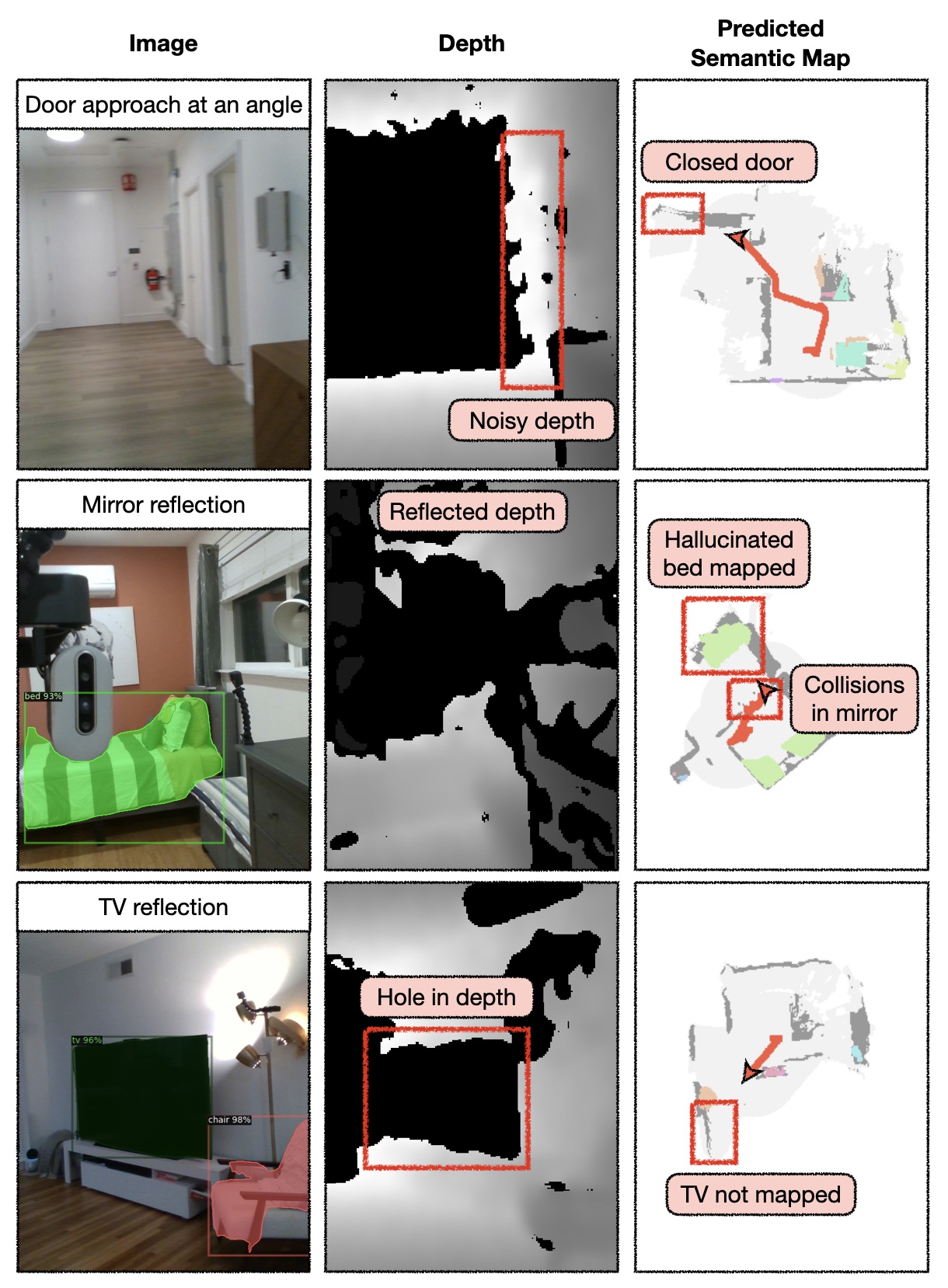}
    \caption{\textbf{Real-world depth sensor error modes.} Top to bottom: \textbf{(A)} Depth noise walls a door when approaching at an angle. \textbf{(B)} Reflection in a mirror creates a duplicate bed on the map instead of an obstacle wall. \textbf{(C)} Reflection on a TV causes depth sensed beyond the sensor limit, and the TV is not mapped.}
    \label{fig:depth_sensor_errors}
\end{figure}

\begin{figure}[h]
    \thisfloatpagestyle{empty}
    \centering
    \includegraphics[height=16cm,keepaspectratio]{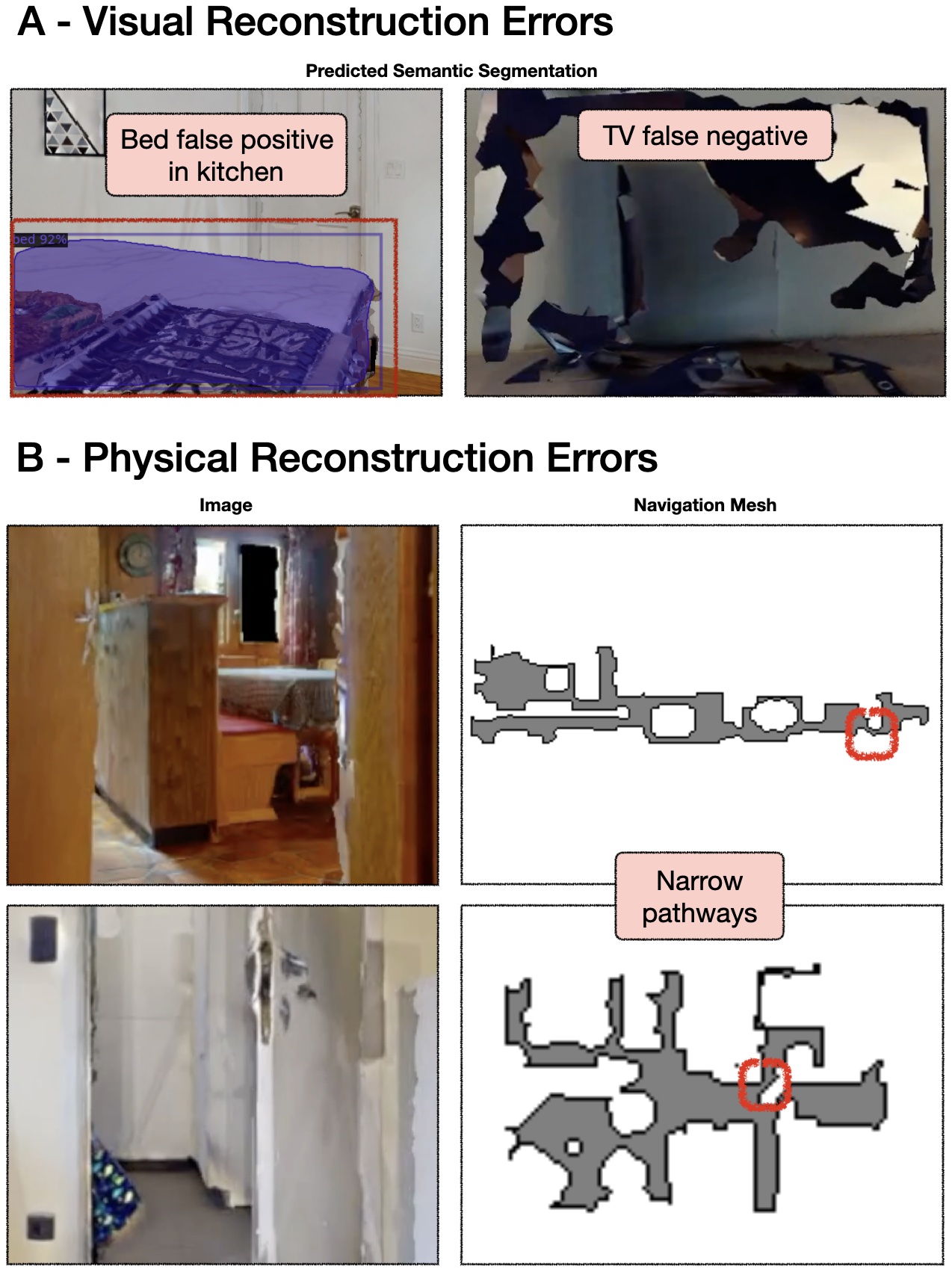}
    \caption{\textbf{Simulation visual and physical reconstruction error modes.} \textbf{(A)} Visual reconstruction errors due to imperfect 3D scanning make segmentation errors more common in sim than reality. \textbf{(B)} Physical reconstruction errors lead to noisy navigation meshes with narrow pathways that are hard to navigate for discrete planners that work well in the real world.}
    \label{fig:reconstruction_errors}
\end{figure}

\begin{figure}[h]
    \centering
    \includegraphics[height=7cm,keepaspectratio]{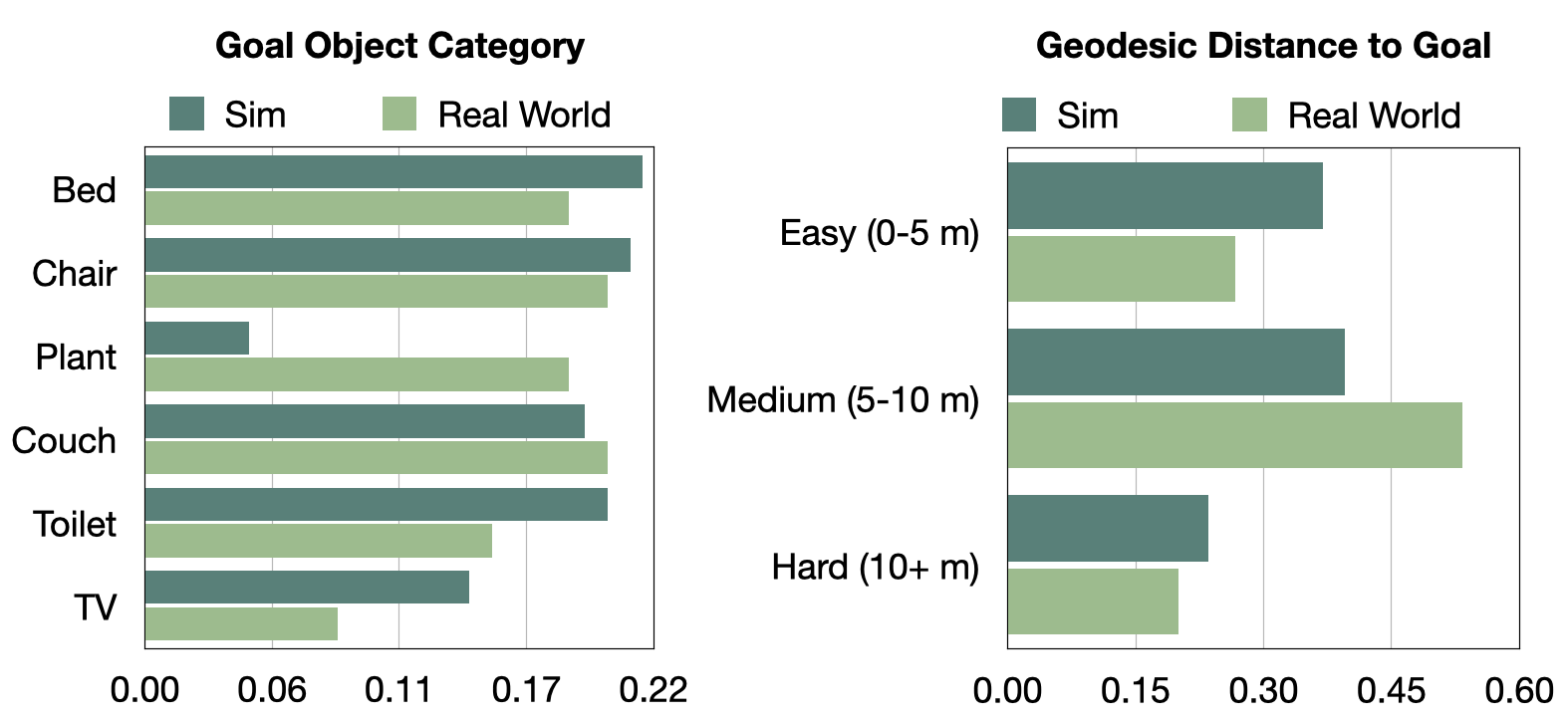}
    \caption{\textbf{Distributions of goal objects and geodesic distances to goal in reality vs. sim.} \textbf{(A)} Distributions of goal objects and geodesic distances to the goal are comparable between a sim benchmark (val set of Habitat Challenge 2022) and our real-world evaluation setting.}
    \label{fig:trajectory_statistics}
\end{figure}

\clearpage
\begin{table*}[]
\tiny
\centering
\caption{
    \small
    \textbf{Navigation performance on all 60 episodes.} \textbf{(A)} Details of navigation performance of classical, modular learning, and end-to-end learning approaches on 60 episodes across 6 object goal categories (“chair”, “couch”, “potted plant”, “toilet”, “tv”, and “bed”) in 6 homes.
}
\label{tab:real_world_results_details}
\begin{tabular}{
>{\columncolor[HTML]{FFFFFF}}c |
>{\columncolor[HTML]{FFFFFF}}c 
>{\columncolor[HTML]{FFFFFF}}c |
>{\columncolor[HTML]{D9EAD3}}c 
>{\columncolor[HTML]{D9EAD3}}r |
>{\columncolor[HTML]{D9EAD3}}c 
>{\columncolor[HTML]{D9EAD3}}r |
>{\columncolor[HTML]{F4CCCC}}c 
>{\columncolor[HTML]{F4CCCC}}r }
\toprule \hline
\multicolumn{3}{c|}{\cellcolor[HTML]{FFFFFF}\textbf{Episode}}                                                                                    & \multicolumn{2}{c|}{\cellcolor[HTML]{FFFFFF}\textbf{Classical}}                                                                                                                                    & \multicolumn{2}{c|}{\cellcolor[HTML]{FFFFFF}\textbf{Modular Learning}}                                                                                                                              & \multicolumn{2}{c}{\cellcolor[HTML]{FFFFFF}\textbf{End-to-end Learning}}                                                                                                                \\ \hline
\textbf{Home}                                & \textbf{Goal Object} & \textbf{\begin{tabular}[c]{@{}c@{}}Shortest Path\\ Length (m)\end{tabular}} & \cellcolor[HTML]{FFFFFF}\textbf{\begin{tabular}[c]{@{}c@{}}Success\\ (SPL / Failure)\end{tabular}} & \cellcolor[HTML]{FFFFFF}\textbf{\begin{tabular}[c]{@{}c@{}}Collisions\\ / Steps\end{tabular}} & \cellcolor[HTML]{FFFFFF}\textbf{\begin{tabular}[c]{@{}c@{}}Success \\ (SPL / Failure)\end{tabular}} & \cellcolor[HTML]{FFFFFF}\textbf{\begin{tabular}[c]{@{}c@{}}Collisions\\ / Steps\end{tabular}} & \cellcolor[HTML]{FFFFFF}\textbf{\begin{tabular}[c]{@{}c@{}}Success\\ (SPL)\end{tabular}} & \cellcolor[HTML]{FFFFFF}\textbf{\begin{tabular}[c]{@{}c@{}}Collisions\\ / Steps\end{tabular}} \\
\toprule \hline
\cellcolor[HTML]{FFFFFF}                     & couch2               & 3.9                                                                         & \cmark$~$ (0.64)                                                                                         & 2 / 51                                                                                        & \cmark$~$ (0.75)                                                                                          & 1 / 39                                                                                        & \cellcolor[HTML]{D9EAD3}\cmark$~$ (0.76)                                                       & \cellcolor[HTML]{D9EAD3}2 / 40                                                                \\
\cellcolor[HTML]{FFFFFF}                     & chair2               & 4.8                                                                         & \cmark$~$ (0.98)                                                                                         & 0 / 35                                                                                        & \cmark$~$ (0.70)                                                                                          & 0 / 45                                                                                        & \xmark$~$                                                                                      & 4 / 44                                                                                        \\
\cellcolor[HTML]{FFFFFF}                     & plant2               & 6.3                                                                         & \cmark$~$ (0.77)                                                                                         & 8 / 63                                                                                        & \cmark$~$ (0.72)                                                                                          & 1 / 64                                                                                        & \xmark$~$                                                                                      & 31 / 200                                                                                      \\
\cellcolor[HTML]{FFFFFF}                     & chair1               & 6.7                                                                         & \cmark$~$ (0.48)                                                                                         & 9 / 114                                                                                       & \cellcolor[HTML]{F4CCCC}\xmark$~$ (TV reflection)                                                            & \cellcolor[HTML]{F4CCCC}25 / 145                                                              & \xmark$~$                                                                                      & 10 / 66                                                                                       \\
\cellcolor[HTML]{FFFFFF}                     & tv1                  & 7.5                                                                         & \cellcolor[HTML]{F4CCCC}\xmark$~$ (TV reflection)                                                        & \cellcolor[HTML]{F4CCCC}15 / 200                                                              & \cmark$~$ (0.42)                                                                                          & 8 / 125                                                                                       & \xmark$~$                                                                                      & 3 / 200                                                                                       \\
\cellcolor[HTML]{FFFFFF}                     & couch1               & 8.1                                                                         & \cmark$~$ (0.33)                                                                                         & 10 / 181                                                                                      & \cmark$~$ (0.74)                                                                                          & 2 / 78                                                                                        & \xmark$~$                                                                                      & 22 / 121                                                                                      \\
\cellcolor[HTML]{FFFFFF}                     & plant1               & 9                                                                           & \cmark$~$ (0.60)                                                                                         & 4 / 130                                                                                       & \cmark$~$ (0.76)                                                                                          & 6 / 127                                                                                       & \cellcolor[HTML]{D9EAD3}\cmark$~$ (0.99)                                                       & \cellcolor[HTML]{D9EAD3}1 / 46                                                                \\
\cellcolor[HTML]{FFFFFF}                     & toilet2              & 10.6                                                                        & \cmark$~$ (0.95)                                                                                         & 3 / 72                                                                                        & \cmark$~$ (0.66)                                                                                          & 6 / 127                                                                                       & \xmark$~$                                                                                      & 16 / 200                                                                                      \\
\cellcolor[HTML]{FFFFFF}                     & toilet1              & 13.3                                                                        & \cmark$~$ (0.89)                                                                                         & 3 / 108                                                                                       & \cmark$~$ (0.77)                                                                                          & 7 / 122                                                                                       & \xmark$~$                                                                                      & 14 / 188                                                                                      \\
\multirow{-10}{*}{\cellcolor[HTML]{FFFFFF}\textbf{1}} & bed1                 & 14.2                                                                        & \cmark$~$ (0.80)                                                                                         & 2 / 125                                                                                       & \cmark$~$ (0.80)                                                                                          & 7 / 135                                                                                       & \cellcolor[HTML]{D9EAD3}\cmark$~$ (0.51)                                                       & \cellcolor[HTML]{D9EAD3}16 / 190                                                              \\
\hline
\cellcolor[HTML]{FFFFFF}                     & toilet2              & 3.5                                                                         & \cellcolor[HTML]{F4CCCC}\xmark$~$ (mirror reflection)                                                    & \cellcolor[HTML]{F4CCCC}37 / 126                                                              & \cmark$~$ (0.49)                                                                                          & 3 / 60                                                                                        & \xmark$~$                                                                                      & 18 / 180                                                                                      \\
\cellcolor[HTML]{FFFFFF}                     & tv1                  & 4.2                                                                         & \cellcolor[HTML]{F4CCCC}\xmark$~$ (segmentation error)                                                   & \cellcolor[HTML]{F4CCCC}7 / 136                                                               & \cmark$~$ (0.99)                                                                                          & 5 / 181                                                                                       & \xmark$~$                                                                                      & 4 / 203                                                                                       \\
\cellcolor[HTML]{FFFFFF}                     & chair3               & 5.5                                                                         & \cmark$~$ (0.64)                                                                                         & 1 / 66                                                                                        & \cmark$~$ (0.97)                                                                                          & 0 / 43                                                                                        & \cellcolor[HTML]{D9EAD3}\cmark$~$ (0.49)                                                       & \cellcolor[HTML]{D9EAD3}5 / 68                                                                \\
\cellcolor[HTML]{FFFFFF}                     & plant2               & 7.7                                                                         & \cmark$~$ (0.49)                                                                                         & 3 / 125                                                                                       & \cmark$~$ (0.63)                                                                                          & 4 / 86                                                                                        & \xmark$~$                                                                                      & 45 / 138                                                                                      \\
\cellcolor[HTML]{FFFFFF}                     & couch2               & 7.8                                                                         & \cmark$~$ (0.35)                                                                                         & 8 / 156                                                                                       & \cmark$~$ (0.66)                                                                                          & 4 / 101                                                                                       & \xmark$~$                                                                                      & 11 / 200                                                                                      \\
\cellcolor[HTML]{FFFFFF}                     & chair1               & 8.2                                                                         & \cmark$~$ (0.86)                                                                                         & 3 / 79                                                                                        & \cmark$~$ (0.79)                                                                                          & 1 / 83                                                                                        & \xmark$~$                                                                                      & 11 / 200                                                                                      \\
\cellcolor[HTML]{FFFFFF}                     & chair2               & 8.6                                                                         & \cmark$~$ (0.64)                                                                                         & 12 / 103                                                                                      & \cmark$~$ (0.47)                                                                                          & 4 / 122                                                                                       & \xmark$~$                                                                                      & 4 / 104                                                                                       \\
\cellcolor[HTML]{FFFFFF}                     & bed1                 & 8.7                                                                         & \cmark$~$ (0.73)                                                                                         & 7 / 75                                                                                        & \cmark$~$ (0.67)                                                                                          & 7 / 103                                                                                       & \xmark$~$                                                                                      & 22 / 189                                                                                      \\
\cellcolor[HTML]{FFFFFF}                     & toilet1              & 9.1                                                                         & \cellcolor[HTML]{F4CCCC}\xmark$~$ (exploration failure)                                                          & \cellcolor[HTML]{F4CCCC}7 / 200                                                               & \cellcolor[HTML]{F4CCCC}\xmark$~$ (depth noise)                                                           & \cellcolor[HTML]{F4CCCC}11 / 137                                                              & \xmark$~$                                                                                      & 1 / 98                                                                                        \\
\cellcolor[HTML]{FFFFFF}                     & couch1               & 9.7                                                                         & \cmark$~$ (0.82)                                                                                         & 2 / 90                                                                                        & \cmark$~$ (0.74)                                                                                          & 5 / 87                                                                                        & \xmark$~$                                                                                      & 30 / 200                                                                                      \\
\multirow{-11}{*}{\cellcolor[HTML]{FFFFFF}\textbf{2}} & plant1               & 9.8                                                                         & \cmark$~$ (0.51)                                                                                         & 3 / 153                                                                                       & \cmark$~$ (0.69)                                                                                          & 2 / 94                                                                                        & \cellcolor[HTML]{D9EAD3}\cmark$~$ (0.51)                                                       & \cellcolor[HTML]{D9EAD3}12 / 153                                                              \\
\hline
\cellcolor[HTML]{FFFFFF}                     & plant2               & 3.1                                                                         & \cmark$~$ (0.76)                                                                                         & 2 / 27                                                                                        & \cmark$~$ (0.71)                                                                                          & 1 / 28                                                                                        & \xmark$~$                                                                                      & 22 / 200                                                                                      \\
\cellcolor[HTML]{FFFFFF}                     & tv1                  & 4.2                                                                         & \cmark$~$ (0.71)                                                                                         & 6 / 68                                                                                        & \cmark$~$ (0.79)                                                                                          & 6 / 41                                                                                        & \xmark$~$                                                                                      & 25 / 173                                                                                      \\
\cellcolor[HTML]{FFFFFF}                     & bed1                 & 4.5                                                                         & \cmark$~$ (0.72)                                                                                         & 1 / 33                                                                                        & \cmark$~$ (0.72)                                                                                          & 1 / 33                                                                                        & \cellcolor[HTML]{D9EAD3}\cmark$~$ (0.27)                                                       & \cellcolor[HTML]{D9EAD3}3 / 109                                                               \\
\cellcolor[HTML]{FFFFFF}                     & bed2                 & 5.2                                                                         & \cmark$~$ (0.49)                                                                                         & 3 / 71                                                                                        & \cmark$~$ (0.60)                                                                                          & 3 / 69                                                                                        & \xmark$~$                                                                                      & 4 / 39                                                                                        \\
\cellcolor[HTML]{FFFFFF}                     & chair1               & 5.9                                                                         & \cmark$~$ (0.99)                                                                                         & 3 / 53                                                                                        & \cmark$~$ (0.84)                                                                                          & 2 / 54                                                                                        & \xmark$~$                                                                                      & 24 / 200                                                                                      \\
\cellcolor[HTML]{FFFFFF}                     & couch1               & 9.3                                                                         & \cmark$~$ (0.69)                                                                                         & 4 / 149                                                                                       & \cmark$~$ (0.45)                                                                                          & 8 / 160                                                                                       & \xmark$~$                                                                                      & 10 / 137                                                                                      \\
\cellcolor[HTML]{FFFFFF}                     & plant1               & 10.2                                                                        & \cmark$~$ (0.46)                                                                                         & 8 / 159                                                                                       & \cmark$~$ (0.60)                                                                                          & 4 / 121                                                                                       & \xmark$~$                                                                                      & 8 / 200                                                                                       \\
\multirow{-8}{*}{\cellcolor[HTML]{FFFFFF}\textbf{3}}  & toilet1              & 10.3                                                                        & \cmark$~$ (0.83)                                                                                         & 11 / 179                                                                                      & \cellcolor[HTML]{F4CCCC}\xmark$~$ (depth noise)                                                           & \cellcolor[HTML]{F4CCCC}8 / 200                                                               & \xmark$~$                                                                                      & 23 / 200                                                                                      \\
\hline
\cellcolor[HTML]{FFFFFF}                     & couch2               & 2                                                                           & \cmark$~$ (0.88)                                                                                         & 0 / 19                                                                                        & \cmark$~$ (0.99)                                                                                          & 0 / 15                                                                                        & \xmark$~$                                                                                      & 32 / 94                                                                                       \\
\cellcolor[HTML]{FFFFFF}                     & bed1                 & 5.5                                                                         & \cellcolor[HTML]{F4CCCC}\xmark$~$ (segmentation error)                                                   & \cellcolor[HTML]{F4CCCC}6 / 133                                                               & \cmark$~$ (0.39)                                                                                          & 2 / 104                                                                                       & \cellcolor[HTML]{D9EAD3}\cmark$~$ (0.60)                                                       & \cellcolor[HTML]{D9EAD3}2 / 68                                                                \\
\cellcolor[HTML]{FFFFFF}                     & plant2               & 5.6                                                                         & \cellcolor[HTML]{F4CCCC}\xmark$~$ (depth noise)                                                          & \cellcolor[HTML]{F4CCCC}4 / 81                                                                & \cmark$~$ (0.20)                                                                                          & 10 / 187                                                                                      & \xmark$~$                                                                                      & 24 / 163                                                                                      \\
\cellcolor[HTML]{FFFFFF}                     & couch1               & 6.1                                                                         & \cmark$~$ (0.90)                                                                                         & 8 / 65                                                                                        & \cmark$~$ (0.28)                                                                                          & 6 / 154                                                                                       & \xmark$~$                                                                                      & 25 / 200                                                                                      \\
\cellcolor[HTML]{FFFFFF}                     & bed2                 & 6.4                                                                         & \cellcolor[HTML]{F4CCCC}\xmark$~$ (segmentation error)                                                   & \cellcolor[HTML]{F4CCCC}22 / 148                                                              & \cmark$~$ (0.61)                                                                                          & 3 / 82                                                                                        & \xmark$~$                                                                                      & 9 / 94                                                                                        \\
\cellcolor[HTML]{FFFFFF}                     & chair1               & 7.2                                                                         & \cmark$~$ (0.86)                                                                                         & 3 / 61                                                                                        & \cmark$~$ (0.81)                                                                                          & 1 / 63                                                                                        & \cellcolor[HTML]{D9EAD3}\cmark$~$ (0.31)                                                       & \cellcolor[HTML]{D9EAD3}11 / 179                                                              \\
\cellcolor[HTML]{FFFFFF}                     & chair2               & 7.3                                                                         & \cmark$~$ (0.82)                                                                                         & 13 / 94                                                                                       & \cmark$~$ (0.72)                                                                                          & 6 / 69                                                                                        & \xmark$~$                                                                                      & 21 / 200                                                                                      \\
\cellcolor[HTML]{FFFFFF}                     & plant1               & 8.3                                                                         & \cmark$~$ (0.70)                                                                                         & 8 / 93                                                                                        & \cmark$~$ (0.62)                                                                                          & 8 / 104                                                                                       & \xmark$~$                                                                                      & 7 / 200                                                                                       \\
\cellcolor[HTML]{FFFFFF}                     & tv1                  & 9.7                                                                         & \cellcolor[HTML]{F4CCCC}\xmark$~$ (TV reflection)                                                        & \cellcolor[HTML]{F4CCCC}5 / 187                                                               & \cellcolor[HTML]{F4CCCC}\xmark$~$ (TV reflection)                                                         & \cellcolor[HTML]{F4CCCC}30 / 188                                                              & \xmark$~$                                                                                      & 43 / 200                                                                                      \\
\cellcolor[HTML]{FFFFFF}                     & plant3               & 10.2                                                                        & \cellcolor[HTML]{F4CCCC}\xmark$~$ (depth noise)                                                          & \cellcolor[HTML]{F4CCCC}15 / 163                                                              & \cellcolor[HTML]{F4CCCC}\xmark$~$ (depth noise)                                                           & \cellcolor[HTML]{F4CCCC}6 / 101                                                               & \xmark$~$                                                                                      & 10 / 81                                                                                       \\
\multirow{-11}{*}{\cellcolor[HTML]{FFFFFF}\textbf{4}} & bed3                 & 14.3                                                                        & \cmark$~$ (0.52)                                                                                         & 4 / 152                                                                                       & \cmark$~$ (0.90)                                                                                          & 2 / 98                                                                                        & \cellcolor[HTML]{D9EAD3}\cmark$~$ (0.85)                                                       & \cellcolor[HTML]{D9EAD3}12 / 101                                                              \\
\hline
\cellcolor[HTML]{FFFFFF}                     & tv1                  & 2                                                                           & \cmark$~$ (0.98)                                                                                         & 1 / 17                                                                                        & \cmark$~$ (0.89)                                                                                          & 0 / 22                                                                                        & \cellcolor[HTML]{D9EAD3}\cmark$~$ (0.90)                                                       & \cellcolor[HTML]{D9EAD3}1 / 19                                                                \\
\cellcolor[HTML]{FFFFFF}                     & chair1               & 3.4                                                                         & \cmark$~$ (0.87)                                                                                         & 1 / 27                                                                                        & \cmark$~$ (0.97)                                                                                          & 1 / 25                                                                                        & \cellcolor[HTML]{D9EAD3}\cmark$~$ (0.83)                                                       & \cellcolor[HTML]{D9EAD3}8 / 34                                                                \\
\cellcolor[HTML]{FFFFFF}                     & couch2               & 4.4                                                                         & \cmark$~$ (0.78)                                                                                         & 0 / 36                                                                                        & \cmark$~$ (0.82)                                                                                          & 0 / 31                                                                                        & \xmark$~$                                                                                      & 9 / 45                                                                                        \\
\cellcolor[HTML]{FFFFFF}                     & plant1               & 4.8                                                                         & \cmark$~$ (0.69)                                                                                         & 2 / 59                                                                                        & \cmark$~$ (0.96)                                                                                          & 1 / 43                                                                                        & \xmark$~$                                                                                      & 16 / 79                                                                                       \\
\cellcolor[HTML]{FFFFFF}                     & chair2               & 5                                                                           & \cmark$~$ (0.99)                                                                                         & 0 / 33                                                                                        & \cmark$~$ (0.99)                                                                                          & 0 / 33                                                                                        & \xmark$~$                                                                                      & 14 / 98                                                                                       \\
\cellcolor[HTML]{FFFFFF}                     & couch1               & 5.2                                                                         & \cmark$~$ (0.57)                                                                                         & 4 / 71                                                                                        & \cmark$~$ (0.77)                                                                                          & 3/ 51                                                                                         & \xmark$~$                                                                                      & 19 / 134                                                                                      \\
\cellcolor[HTML]{FFFFFF}                     & bed2                 & 7.2                                                                         & \cmark$~$ (0.96)                                                                                         & 3 / 58                                                                                        & \cmark$~$ (0.92)                                                                                          & 2 / 62                                                                                        & \cellcolor[HTML]{D9EAD3}\cmark$~$ (0.95)                                                       & \cellcolor[HTML]{D9EAD3}10 / 60                                                               \\
\cellcolor[HTML]{FFFFFF}                     & toilet1              & 8                                                                           & \cellcolor[HTML]{F4CCCC}\xmark$~$ (exploration failure)                                                  & \cellcolor[HTML]{F4CCCC}10 / 200                                                              & \cmark$~$ (0.86)                                                                                          & 3 / 89                                                                                        & \xmark$~$                                                                                      & 26 / 200                                                                                      \\
\cellcolor[HTML]{FFFFFF}                     & bed1                 & 8.7                                                                         & \cmark$~$ (0.82)                                                                                         & 5 / 82                                                                                        & \cmark$~$ (0.37)                                                                                          & 6 / 163                                                                                       & \xmark$~$                                                                                      & 5 / 82                                                                                        \\
\multirow{-10}{*}{\cellcolor[HTML]{FFFFFF}\textbf{5}} & toilet2              & 12.1                                                                        & \cellcolor[HTML]{F4CCCC}\xmark$~$ (exploration failure)                                                  & \cellcolor[HTML]{F4CCCC}10 / 200                                                              & \cellcolor[HTML]{F4CCCC}\xmark$~$ (exploration failure)                                                   & \cellcolor[HTML]{F4CCCC}8 / 200                                                               & \xmark$~$                                                                                      & 17  200                                                                                       \\
\hline
\cellcolor[HTML]{FFFFFF}                     & toilet1              & 3.3                                                                         & \cmark$~$ (0.84)                                                                                         & 2 / 29                                                                                        & \cmark$~$ (0.90)                                                                                          & 1 / 29                                                                                        & \cellcolor[HTML]{D9EAD3}\cmark$~$ (0.69)                                                       & \cellcolor[HTML]{D9EAD3}6 / 39                                                                \\
\cellcolor[HTML]{FFFFFF}                     & chair1               & 3.8                                                                         & \cmark$~$ (0.98)                                                                                         & 1 / 32                                                                                        & \cmark$~$ (0.92)                                                                                          & 1 / 32                                                                                        & \cellcolor[HTML]{D9EAD3}\cmark$~$ (0.78)                                                       & \cellcolor[HTML]{D9EAD3}7 / 45                                                                \\
\cellcolor[HTML]{FFFFFF}                     & chair2               & 4.7                                                                         & \cmark$~$ (0.48)                                                                                         & 3 / 64                                                                                        & \cmark$~$ (0.94)                                                                                          & 1 / 32                                                                                        & \xmark$~$                                                                                      & 12 / 87                                                               \\
\cellcolor[HTML]{FFFFFF}                     & plant1               & 4.8                                                                         & \cmark$~$ (0.55)                                                                                         & 2 / 60                                                                                        & \cmark$~$ (0.36)                                                                                          & 4 / 104                                                                                       & \xmark$~$                                                                                      & 9 / 70                                                                \\
\cellcolor[HTML]{FFFFFF}                     & couch3               & 5.1                                                                         & \cmark$~$ (0.61)                                                                                         & 5 / 76                                                                                        & \cmark$~$ (0.61)                                                                                          & 2 / 60                                                                                        & \xmark$~$                                                                                      & 18 / 149                                                              \\
\cellcolor[HTML]{FFFFFF}                     & couch2               & 7.4                                                                         & \cmark$~$ (0.54)                                                                                         & 4 / 112                                                                                       & \cmark$~$ (0.79)                                                                                          & 4 / 85                                                                                        & \xmark$~$                                                                                      & 16 / 122                                                              \\
\cellcolor[HTML]{FFFFFF}                     & bed1                 & 10.1                                                                        & \cmark$~$ (0.86)                                                                                         & 1 / 94                                                                                        & \cmark$~$ (0.83)                                                                                          & 1 / 83                                                                                        & \xmark$~$                                                                                      & 33 / 200                                                              \\
\cellcolor[HTML]{FFFFFF}                     & couch1               & 10.2                                                                        & \cmark$~$ (0.66)                                                                                         & 1 / 125                                                                                       & \cmark$~$ (0.53)                                                                                          & 6 / 148                                                                                       & \xmark$~$                                                                                      & 24 / 200                                                              \\
\cellcolor[HTML]{FFFFFF}                     & toilet2              & 10.9                                                                        & \cmark$~$ (0.58)                                                                                         & 4 / 139                                                                                       & \cmark$~$ (0.59)                                                                                          & 6 / 187                                                                                       & \xmark$~$                                                                                      & 4 / 139                                                               \\
\multirow{-10}{*}{\cellcolor[HTML]{FFFFFF}\textbf{6}} & bed2                 & 13.1                                                                        & \cellcolor[HTML]{F4CCCC}\xmark$~$ (depth noise)                                                          & \cellcolor[HTML]{F4CCCC}4 / 200                                                               & \cmark$~$ (0.45)                                                                                          & 14 / 186                                                                                      & \xmark$~$                                                                                      & 26 / 200                                                              \\
\hline \bottomrule
\end{tabular}
\end{table*}

\begin{table*}[]
\small
\centering
\caption{
    \small
    \textbf{Navigation performance aggregated by home and goal object.} We compare approaches by Success Rate (SR) and Success weighted by Path Length (SPL). \textbf{(A)} Results are consistent across homes. \textbf{(B)} Results are consistent across goal objects.
}
\label{tab:real_world_results_aggregated}
\begin{tabular}{cccc}
\toprule \hline
\multicolumn{1}{c|}{\textbf{Home}} & \textbf{\begin{tabular}[c]{@{}c@{}}Modular Learning \\ SR (SPL)\end{tabular}} & \textbf{\begin{tabular}[c]{@{}c@{}}Classical \\ SR (SPL)\end{tabular}} & \textbf{\begin{tabular}[c]{@{}c@{}}End-to-end Learning \\ SR (SPL)\end{tabular}} \\
\midrule \hline
\multicolumn{1}{c|}{\textbf{1}}             & \textbf{0.90 (0.63)}                                                          & \textbf{0.90 (0.64)}                                                   & 0.30 (0.23)                                                                      \\
\multicolumn{1}{c|}{\textbf{2}}              & \textbf{0.90 (0.61)}                                                          & 0.70 (0.44)                                                            & 0.10 (0.05)                                                                      \\
\multicolumn{1}{c|}{\textbf{3}}              & 0.88 (0.59)                                                                   & \textbf{1.00 (0.70)}                                                   & 0.13 (0.03)                                                                      \\
\multicolumn{1}{c|}{\textbf{4}}              & \textbf{0.80 (0.49)}                                                          & 0.50 (0.39)                                                            & 0.30 (0.17)                                                                      \\
\multicolumn{1}{c|}{\textbf{5}}              & \textbf{0.90 (0.76)}                                                          & 0.80 (0.67)                                                            & 0.30 (0.27)                                                                      \\
\multicolumn{1}{c|}{\textbf{6}}              & \textbf{1.00 (0.69)}                                                          & 0.90 (0.61)                                                            & 0.20 (0.15)                                                                      \\
\hline \bottomrule \\
\toprule \hline
\multicolumn{1}{c|}{\textbf{Goal}} & \textbf{\begin{tabular}[c]{@{}c@{}}Modular Learning \\ SR (SPL)\end{tabular}} & \textbf{\begin{tabular}[c]{@{}c@{}}Classical \\ SR (SPL)\end{tabular}} & \textbf{\begin{tabular}[c]{@{}c@{}}End-to-end Learning \\ SR (SPL)\end{tabular}} \\
\midrule \hline
\multicolumn{1}{c|}{\textbf{Couch}}        & \textbf{1.00 (0.72)}                                                          & \textbf{1.00 (0.67)}                                                   & 0.08 (0.06)                                                                      \\
\multicolumn{1}{c|}{\textbf{Chair}}        & 0.90 (0.78)                                                                   & \textbf{1.00 (0.82)}                                                   & 0.33 (0.20)                                                                      \\
\multicolumn{1}{c|}{\textbf{Bed}}          & \textbf{1.00 (0.70)}                                                          & 0.70 (0.56)                                                            & 0.45 (0.29)                                                                      \\
\multicolumn{1}{c|}{\textbf{Plant}}         & \textbf{0.90 (0.58)}                                                          & 0.80 (0.53)                                                            & 0.18 (0.14)                                                                      \\
\multicolumn{1}{c|}{\textbf{Toilet}}        & \textbf{0.66 (0.47)}                                                          & 0.55 (0.45)                                                            & 0.11 (0.08)                                                                      \\
\multicolumn{1}{c|}{\textbf{TV}}            & \textbf{0.83 (0.66)}                                                          & 0.50 (0.42)                                                            & 0.20 (0.18)                                                                      \\
\hline \bottomrule
\end{tabular}
\end{table*}

\clearpage
\begin{table*}[h]
\centering
\small

\caption{
\small
    \textbf{Modular learning real vs. sim error modes.} \textbf{(A)} Real-world errors largely stem from depth sensor errors. \textbf{(B)} Sim errors from reconstruction errors: segmentation errors (more common in sim due to imperfect visual reconstruction) and navigation mesh errors (imperfect physical reconstruction).
}
\label{tab:error_modes}
\begin{tabular}{l|ccc} \toprule\hline
\multicolumn{1}{c|}{\textbf{Error mode}} & \textbf{6 real homes} & \textbf{Sim replica} & \textbf{Sim benchmark} \\ \hline \midrule
\textbf{Segmentation error} & 0 / 6 & 2 / 2 & 10.1\% / 18.6\% \\
\textbf{Navigation mesh error} & - & 0 / 2 & 5.5\% / 18.6\% \\
\textbf{Exploration failure} & 1 / 6 & 0 / 2 & 3.0\% / 18.6\% \\ \hline
\textbf{\begin{tabular}[c]{@{}l@{}}Depth sensor error:\\ 1 - Noise closes door\\ 2 - Reflection (mirror, TV)\end{tabular}} & 5 / 6 & - & - \\\hline\bottomrule
\end{tabular} 
\bigskip

\caption{
\small
    \textbf{Modular learning programmatic error analysis on sim benchmark.} We isolate \textbf{(A)} errors due to multi-floor navigation by comparing performance on all episodes and the subset of episodes with start and goal on the first floor, \textbf{(B)} errors due to segmentation failures by introducing ground-truth segmentation, \textbf{(C)} errors due to exploration failures by introducing a very large time budget ($2000$ steps).
}
\label{tab:programmatic_error_analysis}
\begin{tabular}{l|l|c} \toprule\hline
\textbf{Episode Set}                                                                                                         & \textbf{Evaluation Condition}                                                         & \textbf{Successful Episodes} \\ \hline \midrule
All episodes (2000)                                                                                                          & \begin{tabular}[c]{@{}c@{}}500 steps budget, predicted segmentation\end{tabular}    & 1078 / 2000                  \\ \hline
\multirow{3}{*}{\begin{tabular}[l]{@{}l@{}}Episodes with start and \\ goal on first floor (1093)\end{tabular}} & \begin{tabular}[c]{@{}c@{}}500 steps budget, predicted segmentation\end{tabular}     & 733 / 1093                  \\
                                                                                                                            & \begin{tabular}[c]{@{}c@{}}500 steps budget, ground-truth segmentation\end{tabular}  & 843 / 1093                  \\
                                                                                                                            & \begin{tabular}[c]{@{}c@{}}2000 steps budget, ground-truth segmentation\end{tabular} & 876 / 1093 \\ \hline\bottomrule                 
\end{tabular}
\bigskip

\caption{
\small
\textbf{Modular learning manual analysis of remaining errors on sim benchmark.} We manually classify the error mode for each of the remaining $1093 - 876 = 217$ episodes.
}
\label{tab:manual_error_analysis}
\begin{tabular}{l|c} \toprule\hline
\textbf{Error Mode}                                                                                                                                           & \textbf{Episodes} \\ \hline \midrule
Navmesh/planning error (agent stuck in narrow pathway)                                                                                                        & 60 / 217          \\
\begin{tabular}[c]{@{}c@{}}Annotation error (agent reaches instance of goal category \\ that is not annotated as belonging to the goal category)\end{tabular} & 157 / 217         \\ \hline\bottomrule
\end{tabular}
\bigskip

\caption{
\small
\textbf{Modular learning episode outcomes on sim benchmark.} We aggregate programmatic and manual error analyses among single-floor navigation episodes ($1093$ episodes).
}
\label{tab:error_analysis_episode_outcomes}
\begin{tabular}{l|l|l}
\textbf{Outcome}    & \textbf{Episodes} & \textbf{Proportion} \\ \toprule\hline
Segmentation error  & 110 (843 - 733) & 10.1\%              \\
Navigation mesh error       & 60                & 5.5\%               \\
Exploration failure & 33 (876 - 843)  & 3.0\%               \\ \hline
Success             & 890 (733 + 157) & 81.4\%              \\ \hline\bottomrule
\end{tabular}

\end{table*}

\clearpage
\bibliography{scibib}
\bibliographystyle{science}

\end{document}